\documentclass[10pt,twocolumn,letterpaper]{article}

\usepackage[pagenumbers]{cvpr} 
\usepackage[accsupp]{axessibility}
\usepackage[dvipsnames]{xcolor}

\usepackage{epsfig}
\usepackage{graphicx}
\usepackage{amsmath}
\usepackage{amssymb}
\usepackage{amsthm}
\usepackage{url}
\usepackage{nicefrac}  
\usepackage{enumitem}
\usepackage{multirow}
\usepackage{bbm}
\usepackage{float}
\usepackage{tabularx,colortbl}
\usepackage{graphbox}
\usepackage{caption}
\usepackage{mathtools}
\usepackage{mathrsfs}
\usepackage{booktabs}
\usepackage{diagbox}
\usepackage{footmisc}
\usepackage{algorithm2e}
\usepackage{thm-restate}

\newcommand{\B}{\bfseries}

\newcommand{\x}{\mathbf{x}}
\newcommand{\y}{\mathbf{y}}

\newcommand{\indep}{\rotatebox[origin=c]{90}{$\models$}}
\newcommand{\Vu}{\mathbf{u}}
\newcommand{\Vv}{\mathbf{v}}

\DeclareMathOperator*{\argmin}{arg\,min}

\definecolor{cvprblue}{rgb}{0.21,0.49,0.74}
\usepackage[pagebackref,breaklinks,colorlinks,citecolor=cvprblue]{hyperref}

\def\paperID{9085} 
\def\confName{CVPR}
\def\confYear{2024}

\title{Observation-Guided Diffusion Probabilistic Models}

\author{ 
Junoh Kang\footnotemark[1]~~$^1$ \qquad 
Jinyoung Choi\footnotemark[1]~~$^1$ \qquad 
Sungik Choi$^{3}$ \qquad 
Bohyung Han$^{1,2}$ 
\\
Computer Vision Laboratory, $^1$ECE \& $^2$IPAI, Seoul National University \qquad
$^3$LG AI Research
\\
{\tt\small \{junoh.kang, jin0.choi, bhhan\}@snu.ac.kr, sungik.choi@lgresearch.ai }
}

\begin{document}
\twocolumn[{
\maketitle

\centering
\setlength{\tabcolsep}{0mm}
\renewcommand{\arraystretch}{0}
\scalebox{0.99}{
    \begin{tabular}{c @{\extracolsep{1.2mm}} c @{\extracolsep{0mm}} ccc @{\extracolsep{1.2mm}} c @{\extracolsep{0mm}} ccc}
        & \multicolumn{8}{c}{NFEs} \\
        & 10 & 20 & 50 & 100 & 10 & 20 & 50 & 100 \\
        \vspace{1mm}\\
        \rotatebox{90}{ \hspace{2mm} \small{\shortstack{Baseline + \\ Euler}}} &
        \includegraphics[width=0.12\linewidth]{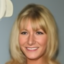} &
        \includegraphics[width=0.12\linewidth]{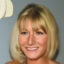} &
        \includegraphics[width=0.12\linewidth]{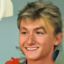} & 
        \includegraphics[width=0.12\linewidth]{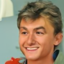} &
        \includegraphics[width=0.12\linewidth]{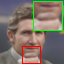} &
        \includegraphics[width=0.12\linewidth]{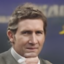} &
        \includegraphics[width=0.12\linewidth]{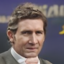} & 
        \includegraphics[width=0.12\linewidth]{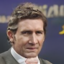}
        \vspace{0.5mm} \\
        \rotatebox{90}{ \hspace{2.6mm} \small{\shortstack{OGDM + \\ Euler}}}&
        \includegraphics[width=0.12\linewidth]{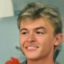} &
        \includegraphics[width=0.12\linewidth]{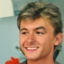} &
        \includegraphics[width=0.12\linewidth]{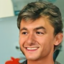} &
        \includegraphics[width=0.12\linewidth]{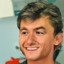} &
        \includegraphics[width=0.12\linewidth]{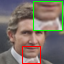} &
        \includegraphics[width=0.12\linewidth]{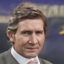} &
        \includegraphics[width=0.12\linewidth]{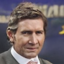} &
        \includegraphics[width=0.12\linewidth]{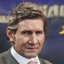}
        \vspace{0.5mm} \\
        \rotatebox{90}{ \hspace{2.6mm} \small{\shortstack{OGDM + \\ S-PNDM}}}&
        \includegraphics[width=0.12\linewidth]{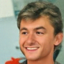} &
        \includegraphics[width=0.12\linewidth]{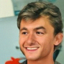} &
        \includegraphics[width=0.12\linewidth]{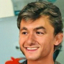} &
        \includegraphics[width=0.12\linewidth]{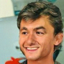} &
        \includegraphics[width=0.12\linewidth]{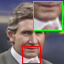} &
        \includegraphics[width=0.12\linewidth]{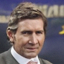} &
        \includegraphics[width=0.12\linewidth]{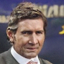} &
        \includegraphics[width=0.12\linewidth]{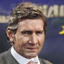}
        \vspace{0.5mm} \\
    \end{tabular}
}
\captionof{figure}{
    Comparisons of images generated by the ADM backbone on the CelebA dataset with deterministic samplers using the same initial noise but different NFEs.
    The entries on the leftmost column of the figure denote the combinations of the training and inference methods.
    (Left) The baseline model generates samples with inconsistent attributes, \eg, gender, hair, \etc, by varying NFEs while our approach preserves such properties.
    (Right) The samples generated by the baseline method with a small number of NFEs tend to be blurry and unrealistic. Also, they have unnaturally bright and textureless areas around the chin of the person.}
\label{fig:teaser}  
\vspace{1cm}
}]

\def\thefootnote{*}\footnotetext{indicates equal contribution.}\def\thefootnote{\arabic{footnote}}


\begin{abstract}
We propose a novel diffusion-based image generation method called the observation-guided diffusion probabilistic model~(OGDM), which effectively addresses the trade-off between quality control and fast sampling.
Our approach reestablishes the training objective by integrating the guidance of the observation process with the Markov chain in a principled way.
This is achieved by introducing an additional loss term derived from the observation based on a conditional discriminator on noise level, which employs a Bernoulli distribution indicating whether its input lies on the (noisy) real manifold or not.
This strategy allows us to optimize the more accurate negative log-likelihood induced in the inference stage especially when the number of function evaluations is limited.
The proposed training scheme is also advantageous even when incorporated only into the fine-tuning process, and it is compatible with various fast inference strategies since our method yields better denoising networks using the exactly the same inference procedure without incurring extra computational cost.
We demonstrate the effectiveness of our training algorithm using diverse inference techniques on strong diffusion model baselines. 
Our implementation is available at \url{https://github.com/Junoh-Kang/OGDM_edm}.%
\end{abstract}


\section{Introduction}
\label{sec:introduction}
Diffusion probabilistic models~\cite{sohl2015deep,ho2020denoising} have shown impressive generation performance in various domains including image~\cite{rombach2022high,dhariwal2021diffusion}, 3D shapes~\cite{zeng2022lion}, point cloud~\cite{luo2021diffusion}, speech~\cite{kong2021diffwave, jeong2021diff-tts}, graph~\cite{niu2020permutation,hoogeboom2022equivariant}, and many others.
The key idea behind these approaches is to formulate data generation as a series of denoising steps of the diffusion process, which sequentially corrupts training data towards a random sample drawn from a prior distribution, \eg, Gaussian distribution.

As diffusion models are trained with an explicit objective, \ie, maximizing log-likelihood, they are advantageous over Generative Adversarial Networks (GANs)~\cite{goodfellow2014generative} in terms of learning stability and sample diversity. 
Moreover, the iterative backward processes and accompanying sampling strategies further improve the quality of samples at the expense of computational efficiency; the tedious inference process involving thousands of network forwarding steps is a critical drawback of diffusion models.

\begin{figure}
    \centering
    \includegraphics[width=0.95\linewidth]{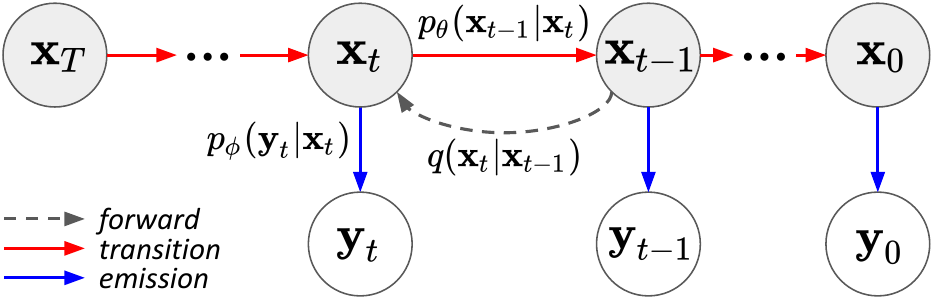} 
    \caption{The graphical model of the proposed denoising process with observations.}
    \label{fig:graph}
    \vspace{-3mm}    
\end{figure}

The step size of diffusion models has a significant impact on the expressiveness of the models as the Gaussian assumption imposed on the reverse (denoising) sampling holds only when the step size is sufficiently small~\cite{sohl2015deep}. 
On the other hand, the backward distribution deviates from the Gaussian assumption as the step size grows, resulting in an inaccurate modeling. 
This exacerbates the discrepancy between the training objective and the negative log-likelihood at inference.
Consequently, performance degradation is inevitable with coarse time steps.
To alleviate the deviation from the true objective caused by large step sizes, our approach incorporates an observation of each state corresponding to the perturbed data.  
To be specific, we consider the data corruption and denoising processes to follow the transition probabilities that respectively align with the forward and backward distributions of DDPM~\cite{ho2020denoising} while an observation at each time step following the emission probability aids in achieving a more accurate backward prediction. 
\cref{fig:graph} depicts the graphical model of our method. 

Our approach offers a significant benefit in the sense that it precisely maximizes the log-likelihood at inference even when employing fast sampling strategies with large step sizes.
The observation process plays an important role during training to adjust the denoising steps towards a more accurate data manifold especially when the reverse process deviates from the Gaussian distribution.
When it comes to the inference stage, the observation process is no longer an accountable factor and hence incurs no additional computational overhead for sampling.

The main question in this approach is what is observable given a state at each time step. 
We define an observation following the Bernoulli distribution on the probability of whether noisy data lies on the manifold of real data with the corresponding noise level.     
From a practical point of view, we implement this observation with a score of a time-dependent discriminator, taking either true denoised samples or fake ones given by the learned denoising network.  

Our main contributions are summarized below:
\vspace{1mm}
\begin{itemize}
    \item We propose an observation-guided diffusion probabilistic model, which accelerates inference procedure while maintaining high sample quality. Our approach only affects the training procedures, resulting in no extra computational or memory overhead during inference. \vspace{1mm}
    \item We derive a principled surrogate loss maximizing the log-likelihood in the observation-guided setting and show its effectiveness in minimizing the KL-divergence between temporally coarse forward and backward processes. \vspace{1mm}
    \item Our training objective is applicable to various inference methods with proper adjustments, which allows us to utilize diverse fast sampling strategies that further improve sample quality. \vspace{1mm}
    \item The proposed technique can be employed for training from scratch or for fine-tuning from pretrained models; compatibility with fine-tuning significantly enhances the practicality of our method.
\end{itemize}
\vspace{1mm}

The rest of this paper is organized as follows.
\cref{sec:related_work} reviews related work and \cref{sec:formulation} describes our main algorithm with the justification of the proposed objective.
We present experimental results and analyses in \cref{sec:results}, discuss future work in \cref{sec:future}, and conclude our paper in \cref{sec:conclusion}.


\section{Related Work}
\label{sec:related_work}
There exists a series of studies on diffusion probabilistic models~\cite{sohl2015deep,ho2020denoising,song2021scorebased} that have contributed to accelerated sampling.
A simple and intuitive method is to simply skip intermediate time steps and sample a subset of the predefined time steps used for training as suggested by DDIM~\cite{song2021denoising}.
By interpreting the diffusion model as solving a specific SDE~\cite{song2021scorebased}, advanced numerical SDE or ODE solvers~\cite{jolicoeur2021gotta,dockhorn2022genie,song2021scorebased,karras2022elucidating,liu2022pseudo} are introduced to speed up the backward process. 
For instance, EDM~\cite{karras2022elucidating} employs a second-order Heun's method~\cite{suli2003introduction} as its ODE solver, demonstrating that simply adopting existing numerical methods can improve performance. 
On the other hand, some papers further refine numerical solvers tailored for diffusion models.
For example, PNDM~\cite{liu2022pseudo} provides a pseudo-numerical solver by combining DDIM and high-order classical numerical methods such as Runge-Kutta~\cite{suli2003introduction} and linear multi-step~\cite{timothy2017numerical}, and GENIE~\cite{dockhorn2022genie} applies a higher-order solver to the DDIM ODE.

On the other hand, \cite{nichol2021improved,bao2022estimating,bao2022analytic,watson2022learning} aim to find the better (optimal) parameters of the reverse process with or without training.
For instance, Analytic-DPM~\cite{bao2022analytic} presents a training-free inference algorithm by estimating the optimal reverse variances under shortened inference steps and computing the KL-divergence between the corresponding forward and reverse processes in analytic forms.
Knowledge distillation~\cite{salimans2022progressive,luhman2021knowledge,meng2023distillation,song2023consistency} is another direction for better optimization, where a single time step in a student model learns to simulate the representations from multiple denoising steps in a teacher model.
Note that our approach is orthogonal and complementary to the aforementioned studies since our goal is to train better denoising networks robust to inference with large step sizes.

There are a few of methods~\cite{xiao2022DDGAN, wang2022diffusion, kim2023refining} that adopt time-dependent discriminators with diffusion process, yet their motivations and intentions differ significantly from ours. 
The time-dependent discriminator in DDGAN~\cite{xiao2022DDGAN} is designed to guide the generator in approximating non-Gaussian reverse processes while Diffusion-GAN~\cite{wang2022diffusion} employs diffusion process to mitigate overfitting of the discriminator.
DG~\cite{kim2023refining}, on the other hand, utilize the discriminator during inference stages to adjust the score estimation additionally.
In contrast, the discriminator in our approach serves as a means to provide observations to the diffusion models during the training phase, without being involved in the inference phase.

\vspace{-1mm}
\section{Observation-Guided Diffusion Probabilistic Models}
\label{sec:formulation}

This section describes the mathematical details of our algorithm and analyzes how to interpret and implement the derived objective function.

\subsection{Properties}
\label{subsec:properties}
The proposed observation-guided diffusion probabilistic model, defined by the graphical model in \cref{fig:graph}, involves two stochastic processes: the state process $\{ \x_t \}_{t=0}^T$ and the observation process $\{ \y_t \}_{t=0}^T$.
The transition and emission probabilities of the forward process, denoted by $q(\x_{t} | \x_{t-1})$ and $q(\y_{t} | \x_{t})$, respectively, are derived by utilizing the following properties given by the graphical model:%
\begin{align}
	\x_{t+1} | \x_{t} \, \indep \, \x_{0:t-1}, \y_{0:t} \quad \text{\&} \quad
	\y_t | \x_t \, \indep \, \x_{0:t-1}, \y_{0:t-1},\label{eq:prop_forward}
\end{align}
where $\indep$ means statistical independence.
In the reverse process, the transition and emission probabilities, $p(\x_{t-1} | \x_{t})$ and $p(\y_{t} | \x_{t})$, are set using the similar properties as
\begin{align}
	\x_{t-1} | \x_t \, \indep \, \x_{T:t+1}, \y_{T:t} \;\; \text{\&} \;\;
	\y_t | \x_t \, \indep \, \x_{T:t+1}, \y_{T:t+1}.\label{eq:prop_backward}
\end{align}
%
\subsection{New surrogate objective}
\label{subsec:extended_surrogate}
Using \cref{eq:prop_forward} and Bayes' theorem, we derive the joint probability of the forward process as follows:
\begin{align}
	q(\x_{1:T}, & \y_{0:T} | \x_0) 
	= q(\x_T | \x_0) \prod_{t=2}^T q(\x_{t-1} | \x_t, \x_0) \prod_{t=0}^T q(\y_{t} | \x_{t}). \label{eq:forward} 
\end{align}
From \cref{eq:prop_backward}, the joint probability of the reverse process is given by
\hspace{-3mm}
\begin{align}
	p(\x_{T:0},\y_{T:0}) 
	= p(\x_T) \prod_{t=T}^{1} p(\x_{t-1}|\x_t) \prod_{t=T}^{0} p(\y_{t} | \x_{t}). \label{eq:backward}
\end{align}
Therefore, we derive the upper bound of the expected negative log-likelihood as 
\begin{align}
	& \mathbb{E}_{\x_0 \sim q} \left[ -\log p(\x_0) \right] \label{eq:nll_train} \\
	&= \mathbb{E}_{\x_0 \sim q} \left[ \log \mathbb{E}_{\x_{1:T}, \y_{0:T} \sim q}  \cfrac{q(\x_{1:T},\y_{0:T} | \x_0)}{p(\x_{0:T},\y_{0:T})} \right] \\
	& \leq \mathbb{E}_{\x_0 \sim q} \mathbb{E}_{\x_{1:T}, \y_{0:T} \sim q} \left[ \log \cfrac{q(\x_{1:T},\y_{0:T} | \x_0)}{p(\x_{0:T},\y_{0:T})} \right] \\
	&= \mathbb{E}_{\x_{0:T}, \y_{0:T} \sim q} \left[ \log \cfrac{q(\x_T|\x_0) \prod_{t=2}^{T} q(\x_{t-1}|\x_t, \x_0)}{p(\x_T) \prod_{t=T}^{1} p(\x_{t-1}|\x_t)} \right] \nonumber \\
	&\quad + \mathbb{E}_{\x_{0:T}, \y_{0:T} \sim q} \left[ \log \cfrac{\prod_{t=0}^{T} q(\y_{t} | \x_{t})}{\prod_{t=T}^{0} p(\y_{t} | \x_{t})} \right] \\
	&= D_\text{KL}(q(\x_T|\x_0)||p(\x_T)) + E_q \left[-\log p(\x_0 |\x_1) \right] \nonumber \\
	&\quad + \sum_{t=2}^{T} D_\text{KL}(q(\x_{t-1}|\x_t,\x_0) || p(\x_{t-1}|\x_t)) \nonumber \\
	&\quad + \sum_{t=0}^{T} D_\text{KL}(q(\y_t|\x_t) || p(\y_t|\x_t)). \label{eq:intermediate objective}
\end{align}
where the first inequality is derived by the Jensen's inequality.

\paragraph{Transition probabilities} 
From \cite{ho2020denoising}, the forward transition probabilities are given by
\begin{align}
	q(\x_0) &\coloneqq \mathrm{P}_\text{data}(\x_0) \quad \text{and} \\
	q(\x_t | \x_{t-1}) &\coloneqq \mathcal{N}(\x_{t};\sqrt{1-\beta_t}\x_{t-1} , \beta_t \mathbf{I}),
	\label{eq:ddpm_forward} 
\end{align}
where $\{\beta_t\}_{t=1}^T$ are predefined constants.
The backward transition probabilities are defined as
\begin{align}
	p_\theta(\x_T) &\coloneqq \mathcal{N}(\x_T; \mathbf{0}, \mathbf{I}) \quad \text{and} \quad \\
	p_\theta(\x_{t-1} | \x_{t}) &\coloneqq \mathcal{N}\hspace{-0.5mm}\left( \x_{t-1};\frac{1}{\sqrt{1-\beta_t}}(\x_{t} + \beta_t s_\theta(\x_t,t)) , \beta_t \mathbf{I} \right), 
	\label{eq:ddpm_backward}
\end{align}
where $s_\theta(\cdot, \cdot)$ denotes a neural network parameterized by $\theta$.
Due to the following equation,
\begin{align}
	& q(\x_{t-1} | \x_{t}, \x_0) \label{eq:ddpm_backward_obj} \\
	&= \mathcal{N} \left(\x_{t-1};\frac{1}{\sqrt{1-\beta_t}}(\x_{t} + \beta_t \nabla\log q(\x_t|\x_0)), \bar{\beta}_t \mathbf{I} \right), \nonumber
\end{align}
where $\bar\alpha_t = \prod_{s=1}^{t}(1-\beta_s)$ and $\bar{\beta}_t = \frac{1-\bar\alpha_{t-1}}{1-\bar\alpha_{t}} \beta_t$, the first three terms of \cref{eq:intermediate objective} are optimized by the following loss function:
\begin{align}
	\sum_{t=1}^{T} \lambda_t ||\epsilon - \epsilon_\theta(\sqrt{\bar\alpha_t} \x_0 + \sqrt{1-\bar\alpha_t}\epsilon , t)||^2_2 + C,  \label{eq:transition loss}
\end{align}
where $\epsilon_\theta(\x_t, t) = \cfrac{s_\theta(\x_t,t)}{\sqrt{1-\bar\alpha_t}}$ and $C$ is a constant.

\paragraph{Emission probabilities}
We interpret the last term in \cref{eq:intermediate objective} as an observation about whether the state, $\x_t$, is on the real data manifold or not.
Then, the emission probability of the forward and backward processes are defined by Bernoulli distributions as
\begin{align}
	q(\y_t | \x_t) \coloneqq \text{Ber}(1) \enspace \text{\&} \enspace p(\y_t | \x_t) \coloneqq \text{Ber}(D(f(\x_t))),
\label{eq:emission_probability}
\end{align}
where $f(\cdot)$ is an arbitrary function that projects an input onto a known manifold and $D(\cdot)$ indicates the probability that an input belongs to the manifold of real data.
Hence, the KL-divergence of emission, \ie, the last term of \cref{eq:intermediate objective}, is redefined via two different Bernoulli distributions in \cref{eq:emission_probability}. 
Eventually, the KL-divergence between two emission distributions is replaced by a log-likelihood of the manifold embedding as follows:
\begin{align}
	\sum_{t=0}^{T}D_\text{KL} & (q(\y_t | \x_t) || p(\y_t | \x_t))
	= \sum_{t=0}^{T}- \log (D(f(\x_t))). \label{eq:emission_loss} 
\end{align}
%

\subsection{Manifold embedding and likelihood function}
\label{subsec:manifold_embedding}

We now discuss a technically feasible way to implement \cref{eq:emission_loss}. 
The only undecided components in \cref{eq:emission_loss} are the projection function to a known manifold, $f(\cdot)$, and the likelihood function, $D(\cdot)$.
We define $f(\cdot)$ as a function projecting $\x_{t}$ onto a manifold of $\x_{t-s} \sim q_{t-s}$  ($t \geq s$).
With the diffusion model, this can be done by running one discretization step of a numerical ODE solver from the noise level of $t$ to $t-s$, denoted by $\Phi(\x_t, t,s; \theta)$.
We implement the projection function using the solver as follows:
\begin{align}
	f_\theta(\x_t) \coloneqq \hat\x_{t-s}^\theta = \Phi(\x_t, t,s; \theta). \label{eq:prediction network}
\end{align}
Note that $s$ is a sample drawn from a uniform distribution, $\mathcal{U}(1,\min(t,\lfloor kT \rfloor ))$, where $k \in [0, 1]$ is the hyperparameter that determines the lookahead range in the backward direction.
We utilize a step of the Euler method\footnote{Refer to \cref{alg:euler_method} in \cref{app:solvers}.\label{fn:euler}} or the Heun's method\footnote{Refer to \cref{alg:heun_method} in \cref{app:solvers}.\label{fn:heun}} to realize the projection function.

In addition, we design $D(\cdot)$ as $D_\phi(\cdot)^\gamma$, a constant power of a discriminator function, $D_\phi(\cdot)$, which distinguishes between projected data from 1) the prediction of the denoising network and 2) real data.
Such a design is motivated by the right-hand side of \cref{eq:emission_loss}, which resembles the objective of the generator in non-saturating GAN~\cite{goodfellow2014generative}; we employ a time-dependent discriminator taking the projected data, $t$ and $s$, as its inputs.
\begin{figure}
    \centering
    \includegraphics[width=0.8\linewidth]{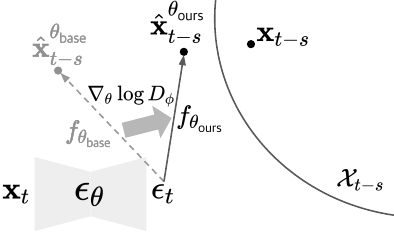} 
    \caption{The role of the discriminator in our objective.
	$\theta_{\text{ours}}$ and $\theta_{\text{base}}$ denote the denoising parameters learned by the proposed method and the baseline, respectively.  
	The proposed training method nudges the prediction of $\hat\x_{t-s}^\theta$ closer to the exact state space than the original.}
    \label{fig:disc_guide}    
\end{figure}
%

\subsection{Training objectives}
\label{subsec:new_objective}

By reformulating the transition and emission probabilities as discussed in \cref{subsec:extended_surrogate,subsec:manifold_embedding}, the first three terms of \cref{eq:intermediate objective} become identical to \cref{eq:transition loss} while the last term is set as $-\gamma \log (D_\phi(\hat\x_{t-s}^\theta,t,s))$.
Therefore, the final training objective of the diffusion model with a network design parametrized by $\theta$ is given by
\begin{align}
	\underset{\theta}{\min}~ 
	\mathbb{E}_{\x_0, \epsilon, t,s} \biggr[ 
	&\underbrace{\lambda_t ||\epsilon - \epsilon_\theta(\sqrt{\bar\alpha_t} \x_0 + \sqrt{1-\bar\alpha_t}\epsilon , t)||^2_2}_{\mathcal{L}_\text{transition}} \nonumber\\
	&- \gamma\underbrace{\log (D_\phi(\hat\x_{t-s}^\theta,t,s))}_{-\mathcal{L}_\text{emission}} \biggr], \label{eq:our_diffusion_objective} 
\end{align}
where $D_\phi(\cdot)$ denotes a discriminator and $\gamma$ is a hyperparameter.

Besides optimizing $\theta$, we need to train $D_\phi(\cdot)$ to distinguish real data from the predictions of the diffusion model.
Following GANs~\cite{goodfellow2014generative}, the training objective is given by
\begin{align}
	\max_{\phi}~
	\mathbb{E}_{\x_0, \epsilon, t, s} 
	[&\log(D_\phi(\x_{t-s},t,s)) \nonumber\\
	&~+ \log(1-D_\phi(\hat\x_{t-s}^\theta,t,s))]. \label{eq:disc_loss}
\end{align}
We perform an alternating optimization, where the two objective functions in \cref{eq:our_diffusion_objective,eq:disc_loss} take turns until convergence. 

For inference, the diffusion model $\epsilon_\theta(\cdot)$ is only taken into account and the discriminator $D_\phi(\cdot)$ is not required.
Therefore, our approach incurs no extra computational overhead for sample generations.


\subsection{Analysis of the observation-induced loss} 
\label{subsec:analysis_final_obj}

We further analyze the surrogate of the negative log-likelihood of a generated sample and explain how the proposed observation-induced loss affects the surrogate.

\subsubsection{Negative log-likelihood at inference}
The negative log-likelihood of a generated sample $\x_0 \sim p_\theta$ is given by 
\begin{align}
	&\mathbb{E}_{\x_0 \sim p_\theta} [ -\log q(\x_0)] \label{eq:nll_sample} \\ 
	&\leq \sum_2^N D_\text{KL}\left(p_\theta(\x_{\tau_{i-1}} | \x_{\tau_{i}}) || q(\x_{\tau_{i-1}} | \x_{\tau_{i}}) \right)  
	+ \mathbb{E}_{p_\theta}[q(\x_{\tau_0}|\x_{\tau_1})], \nonumber
\end{align}
where $\tau_0=0 < \tau_1 < \cdots < \tau_N=T$ is a subsequence of time steps selected for fast sampling and $\x_0$ is sampled from $p_\theta$ unlike \cref{eq:nll_train}.
Here, $p_\theta(\x_{\tau_{i-1}} | \x_{\tau_{i}})$ is defined similarly to \cref{eq:ddpm_backward} by replacing $\beta_t$ with $\tilde\beta_{\tau_i}=1 - \frac{\bar\alpha_{\tau_{i}}}{\bar\alpha_{\tau_{i-1}}}$.%

\subsubsection{Approximation on true reverse distribution}

While $p_\theta(\x_{\tau_{i-1}} | \x_{\tau_{i}})$ takes the tractable form of a Gaussian distribution, the true reverse distribution, $q(\x_{\tau_{i-1}} | \x_{\tau_{i}})$, is still infeasible for estimating the KL-divergence in the right-hand side of \cref{eq:nll_sample}. 
To simulate the true reverse density function with $\tilde\beta_{\tau_{i}}$, we use a weighted geometric mean of its asymptotic distributions corresponding to $\tilde\beta_{\tau_i} \approx 0$ or $1$.

\vspace{0mm}
For notational simplicity, let $\x_{\tau_{i-1}} = \Vu$, $\x_{\tau_t}=\Vv$, and $\tilde\beta_{\tau_i} = \beta$. 
Then, we denote the true reverse density function as $p_{\Vu|\Vv}^{(\beta)}(\Vu|\Vv)$, for $\Vu \sim p_\Vu$ and $\Vv|\Vu \sim \mathcal{N}(\sqrt{1-\beta}\Vu, \beta \mathbf{I})$. 

\begin{restatable}[]{lemma}{limit} \label{lemma1}
	For $\Vu \sim p_\Vu$ and $\Vv|\Vu \sim \mathcal{N}(\sqrt{1-\beta}\Vu, \beta \mathbf{I})$, we obtain the following two asymptotic distributions of $p_{\Vu|\Vv}^{(\beta)}(\Vu|\Vv)$\textup{:}
	\begin{align}
		& p_{\Vu|\Vv}^{(\beta)}(\Vu|\Vv) \approx \mathcal{N}\left(\Vu;\frac{1}{\sqrt{1-\beta}}(\Vv + \beta \nabla\log p_\Vu(\Vv)), \beta \mathbf{I} \right)
		\nonumber\\
		& \hspace{14em}\text{for } 0 < \beta \ll 1, \label{eq:asymptotic1} \\
		& \lim_{\beta \rightarrow 1^-} p_{\Vu|\Vv}^{(\beta)}(\Vu|\Vv) = p_\Vu(\Vu). \label{eq:asymptotic2}	
	\end{align}
\end{restatable}
\begin{proof}
	Refer to \cref{app:lemma1}.
\end{proof}
The weighted geometric mean of the two asymptotic distributions in \cref{eq:asymptotic1,eq:asymptotic2} is given by
\begin{align}
	&q_{\Vu|\Vv}^{(\xi)}(\Vu|\Vv) \label{eq:mixture} \\
	&\coloneqq C_{\xi} ~ \mathcal{N} \left(\Vu;\frac{1}{\sqrt{1-\beta}}(\Vv + \beta \nabla\log p_\Vu(\Vv)), \beta \mathbf{I} \right)^{1-\xi} p_\Vu(\Vu)^{\xi}, \nonumber
\end{align}
where $C_\xi$ is the normalization constant and $\xi$ determines the weight of each component. 
We further define a mapping function $\xi(\beta)$ that minimizes the difference between $p_{\Vu|\Vv}^{(\beta)}(\Vu|\Vv)$ and $q_{\Vu|\Vv}^{(\xi)}(\Vu|\Vv)$ as
\begin{align}
	\xi(\beta) \coloneqq \argmin_{\xi \in [0,1]} \int_{-\infty}^{\infty} \left( q_{\Vu|\Vv}^{(\xi)}(\Vu|\Vv) - p_{\Vu|\Vv}^{(\beta)}(\Vu|\Vv) \right)^2 d\Vu.
\end{align}
The existence of $\xi(\beta)$ is clear under continuity of $q_{\Vu|\Vv}^{(\xi)}(\Vu|\Vv)$ with respect to $\xi$.
For the rest of the analysis, we approximate $p_{\Vu|\Vv}^{(\beta)}(\Vu|\Vv)$ by $q_{\Vu|\Vv}^{(\xi(\beta))}(\Vu|\Vv)$; the validity of the approximation is discussed in \cref{app:proposition1} with more detailed empirical study results.
Finally, by substituting the variables back as $\Vu=\x_{\tau_{i-1}}$, $\Vv=\x_{\tau_i}$, and $\beta = \tilde\beta_{\tau_i}$, the true reverse density function is approximated by
\begin{align}
	&q(\x_{\tau_{i-1}} | \x_{\tau_{i}}) \label{eq:backward_approximation} \\
	&\approx C_{\xi(\tilde\beta_{\tau_i})} \mathcal{N}(\x_{\tau_{i-1}}; \boldsymbol{\mu}_{\tau_i}, \tilde\beta_{\tau_i}\mathbf{I})^{1-\xi(\tilde\beta_{\tau_i})}
	  q(\x_{\tau_{i-1}})^{\xi(\tilde\beta_{\tau_i})}, \nonumber
\end{align}
where $\boldsymbol{\mu}_{\tau_i} = \frac{1}{\sqrt{1-\tilde\beta_{\tau_i}}}(\x_{\tau_i} + \tilde\beta_{\tau_i}\nabla\log q(\x_{\tau_i}))$.

\subsubsection{Interpretation of \cref{eq:nll_sample}}
By using \cref{eq:backward_approximation}, we factorize the KL-divergence term in \cref{eq:nll_sample} into a sum of two KL-divergences as follows:
\begin{align}
	&D_\text{KL}(p_\theta(\x_{\tau_{i-1}} | \x_{\tau_{i}}) || q(\x_{\tau_{i-1}} | \x_{\tau_{i}}))  \nonumber \\
	& \approx (1-\xi(\tilde\beta_{\tau_i})) D_\text{KL}(p_\theta(\x_{\tau_{i-1}} | \x_{\tau_{i}}) ||\mathcal{N}(\x_{\tau_{i-1}}; \boldsymbol{\mu}_{\tau_i}, \tilde\beta_{\tau_i}\mathbf{I})) \nonumber \\
	&\quad + \xi(\tilde\beta_{\tau_i}) D_\text{KL}(p_\theta(\x_{\tau_{i-1}} | \x_{\tau_{i}}) || q(\x_{\tau_{i-1}})) + \log C_{\xi(\tilde\beta_{\tau_i})} \nonumber\\
	&= (1 - \xi(\tilde\beta_{\tau_i})) || s_\theta(\x_{\tau_i}, \tau_i) - \nabla\log q(\x_{\tau_i})||_2^2   \nonumber \\
	&\quad + \xi(\tilde\beta_{\tau_i}) D_\text{KL}(p_\theta(\x_{\tau_{i-1}} | \x_{\tau_{i}}) || q(\x_{\tau_{i-1}})) + C, \label{eq:inference_kl_final} 
\end{align}
where $C$ is a constant.

While $\mathcal{L}_{\text{transition}}$ in~\cref{eq:our_diffusion_objective} minimizes the first term of the last equation in~\cref{eq:inference_kl_final}, $\mathcal{L}_{\text{emission}}$ in~\cref{eq:our_diffusion_objective} minimizes the JS-divergence between two distributions \cite{goodfellow2014generative}. 
Although the JS-divergence has different properties from the KL-divergence, both quantities are minimized when two distributions are equal; the minimization of the JS-divergence effectively reduces the second term of \cref{eq:inference_kl_final} in practice.

Note that the vanilla diffusion models neglect the second term of \cref{eq:inference_kl_final} while DDGAN~\cite{xiao2022DDGAN} disregards the first term of \cref{eq:inference_kl_final}.
On the contrary, the proposed method considers both components, leading to effective optimization.
In practice, both $\xi(\tilde\beta_{\tau_i})$ and $1-\xi(\tilde\beta_{\tau_i})$ are expected to be non-trivial in fast sampling with a relatively large value of $\tilde\beta_{\tau_i}$. 
The behaviors of $\xi(\tilde\beta_{\tau_i})$ are demonstrated in \cref{fig:proposition1}~(a) and (b) in \cref{app:proposition1}.

\begin{table*}[ht]
    \centering
    \caption{
        FID and recall scores for various NFEs when the projection function $f_\theta(\cdot)$ aligns to the sampler as Euler method\footref{fn:euler}.
        `EDM cond.' denotes that the model is trained with class labels.
        `OGDM' represents that the models are trained from scratch while `OGDM (ft)' indicates that the models are fine-tuned from the pretrained baseline models.
    }
    \label{tab:quant_euler}
    \vspace{-1mm}
        \setlength{\tabcolsep}{3.2mm}
    \scalebox{0.9}{
        \begin{tabular}{clcccccccc}
    \toprule
    \multicolumn{2}{c}{\diagbox[height=1.2\line]{}{\hspace{15mm} NFEs}} & 
    \multicolumn{2}{c}{25} & \multicolumn{2}{c}{20} & \multicolumn{2}{c}{15} & \multicolumn{2}{c}{10} \\ 
    \cmidrule(lr){3-4}\cmidrule(lr){5-6}\cmidrule(lr){7-8}\cmidrule(lr){9-10}
    Dataset~(Backbone) & Method & FID$\downarrow$ & Rec.$\uparrow$ & FID$\downarrow$ & Rec.$\uparrow$ & FID$\downarrow$ & Rec.$\uparrow$ & FID$\downarrow$ & Rec.$\uparrow$ \\ \hline
    
    \multirow{3}{*}{CIFAR-10~(ADM)}
    & Baseline & 7.08 & 0.583 & 8.05 & 0.582 & 9.93 & 0.567 & 15.20 & 0.527  \\
    & OGDM & \B 6.26 & \B 0.587 & \B 6.81 & \B 0.587 & \B 7.96 & \B 0.578 & 11.63 & 0.546  \\
    & OGDM (ft) & 6.69 & 0.582 & 7.26 & 0.581 & 8.15 & 0.571 & \B 11.18 & \B 0.549  \\
    \cmidrule(lr){1-10}
    
    \multirow{2}{*}{CIFAR-10~(EDM)} 
    & Baseline & 5.32 & 0.572 & 6.82 & 0.558 & 10.02 & 0.524 & 19.32 & 0.452 \\
    & OGDM (ft) & \B 3.21 & \B 0.603 & \B 3.53 & \B 0.600 & \B 4.64 & \B 0.587 & \B 9.28 & \B 0.546 \\
    \cmidrule(lr){1-10}

    \multirow{2}{*}{CIFAR-10~(EDM cond.)} 
    & Baseline  & 4.95 & 0.567 & 6.23 & 0.546 & 8.81 & 0.514 & 15.57 & 0.434 \\
    & OGDM (ft) & \B 2.56 & \B 0.599 & \B 2.83 & \B 0.589 & \B 3.67 & \B 0.572 & \B 6.85 & \B 0.528 \\
    \cmidrule(lr){1-10}
    
    \multirow{3}{*}{CelebA~(ADM)}
    & Baseline & 7.20 & 0.441 & 7.88 & 0.429 & 9.34 & 0.392 & 11.92 & 0.315 \\
    & OGDM & \B 3.80 & 0.541 & \B 3.94 & 0.534 & 5.06 & 0.502 & 7.91 & 0.451 \\
    & OGDM (ft) & 4.61 & \B 0.576 & 4.61 & \B 0.571 & \B 4.80 & \B 0.552 & \B 7.04 & \B 0.504 \\
    \cmidrule(lr){1-10}
    
    \multirow{2}{*}{LSUN Church~(LDM)}
        & Baseline & 7.87 & 0.443 & 8.40 & 0.434 & 8.83 & 0.399 & 15.02 & 0.326 \\
        & OGDM (ft) & \B 7.46 & \B 0.449 & \B 7.92 & \B 0.444 & \B 8.76 & \B 0.402 & \B 14.84 & \B 0.331 \\
    \bottomrule    
        \end{tabular}
        }
\vspace{3mm}
\end{table*}
\begin{table*}[t]
    \centering
    \caption{
        FID and recall scores for various NFEs when the projection function $f_\theta(\cdot)$ aligns to the sampler as Heun's method\footref{fn:heun}.
        `OGDM (ft)' indicates that the models are fine-tuned from the pretrained baseline models.
    }
    \label{tab:quant_heun}
    \vspace{-1mm}
        \setlength{\tabcolsep}{2.2mm}
    \scalebox{0.9}{
        \begin{tabular}{clcccccccccc}
    \toprule
    \multicolumn{2}{c}{\diagbox[height=1.2\line]{}{\hspace{14mm} NFEs}} & 
    \multicolumn{2}{c}{35} & \multicolumn{2}{c}{25} & \multicolumn{2}{c}{19} & \multicolumn{2}{c}{15} & \multicolumn{2}{c}{11} \\ 
    \cmidrule(lr){3-4}\cmidrule(lr){5-6}\cmidrule(lr){7-8}\cmidrule(lr){9-10}\cmidrule(lr){11-12}
    Dataset~(Backbone) & Method & FID$\downarrow$ & Rec.$\uparrow$ & FID$\downarrow$ & Rec.$\uparrow$ & FID$\downarrow$ & Rec.$\uparrow$ & FID$\downarrow$ & Rec.$\uparrow$ & FID$\downarrow$ & Rec.$\uparrow$ \\ \hline
    
    \multirow{2}{*}{CIFAR-10~(EDM)}
    & Baseline & \B 2.07 & 0.618 & 2.19 & 0.616 & 2.73 & 0.616 & 4.48 & 0.604 & 14.71 & 0.536 \\
    & OGDM (ft) & 2.15 & \B 0.620 & \B 2.17 & \B 0.622 & \B 2.56 & \B 0.620 & \B 4.21 & \B 0.619 & \B 13.54 & \B 0.589 \\
    \bottomrule    
        \end{tabular}
        }
\end{table*}
\section{Experiments}
\label{sec:results}

This section first describes the evaluation protocol in this work, and then presents quantitative and qualitative results of OGDM in comparison to other baselines.

\subsection{Evaluation protocol}
\label{subsec:exp_detail}

The datasets, model architectures, and training and evaluation methods are as follows.

\paragraph{Datasets}
We perform the unconditional image generation experiment on several standard benchmarks with diverse resolutions---CIFAR-10~\cite{krizhevsky2009learning} (32$\times$32), CelebA~\cite{liu2015faceattributes} (64$\times$64) and LSUN Church~\cite{yu2015lsun} (256$\times$256).

\paragraph{Architectures}
We apply our method to three strong baselines, ADM\footnote{https://github.com/openai/guided-diffusion\label{fn:adm}}~\cite{dhariwal2021diffusion} on CIFAR-10 and CelebA, EDM\footnote{https://github.com/NVlabs/edm\label{fn:edm}}~\cite{karras2022elucidating} on CIFAR-10, and LDM\footnote{https://github.com/CompVis/latent-diffusion}~\cite{rombach2022high} on LSUN Church, using their official source codes. 
For the implementation of the time-dependent discriminator, we mostly follow the architecture proposed in Diffusion-GAN~\cite{wang2022diffusion}, which is based on the implementation of StyleGAN2\footnote{https://github.com/NVlabs/stylegan2-ada-pytorch}~\cite{karras2020training} while time indices are injected into the discriminator as in the conditional GAN.
The only modification in our implementation is an additional time index, $s$ in \cref{eq:prediction network}, which denotes the number of lookahead time steps from the current time index, $t$.

\paragraph{Training}
We use the default hyperparameters and optimization settings provided by the official codes of baseline algorithms for all experiments except for discriminator training.
We consistently obtain favorable results with $k\in[0.1, 0.2]$ and $\gamma\in[0.005, 0.025]$ across all datasets and present our choices of the hyperparameters for reproducibility in \cref{app:hyperparameter}.

\paragraph{Evaluation}
We measure FID~\cite{heusel2017gans} and recall~\cite{kynkaanniemi2019improved} using the implementation provided by ADM\footref{fn:adm} for quantitative evaluation.
To compute FID, we use the full training data as a reference set and 50K generated images as an evaluation set.
For the recall metric, we utilize 50K images for both reference and generated sets.

\begin{table*}[t]
    \centering
    \caption{
        FID and recall scores for various NFEs when the projection function $f_\theta(\cdot)$ is a step of Euler method\footref{fn:euler} while two different PNDM~\cite{liu2022pseudo} algorithms are used as samplers.
        `EDM cond.' denotes that the model is trained with class labels.
        `OGDM' represents that the models are trained from scratch while `OGDM (ft)' indicates that the models are fine-tuned from the pretrained baseline models.}
    \label{tab:quant_pndm}
    \vspace{-1mm}
        \setlength{\tabcolsep}{2.2mm}
    \scalebox{0.9}{
        \begin{tabular}{cclcccccccc}
    \toprule
    \multicolumn{3}{c}{\diagbox[height=1.2\line]{}{\hspace{24mm} NFEs}} & 
    \multicolumn{2}{c}{25} & \multicolumn{2}{c}{20} & \multicolumn{2}{c}{15} & \multicolumn{2}{c}{10} \\ 
    \cmidrule(lr){4-5}\cmidrule(lr){6-7}\cmidrule(lr){8-9}\cmidrule(lr){10-11}
    Sampler & Dataset~(Backbone) & Method & FID$\downarrow$ & Rec.$\uparrow$ & FID$\downarrow$ & Rec.$\uparrow$ & FID$\downarrow$ & Rec.$\uparrow$ & FID$\downarrow$ & Rec.$\uparrow$ \\ \hline
    
    \multirow{12}{*}{S-PNDM} & \multirow{2}{*}{CIFAR-10~(ADM)} 
    & Baseline & 5.31 & 0.601 & 5.95 & 0.596 & 7.09 & 0.577 & 10.32 & 0.548 \\
    & & OGDM & \B 5.03 & \B 0.611 & \B 5.09 & \B 0.605 & \B 5.58 & \B 0.601 & \B 7.54 & \B 0.582 \\
    \cmidrule(lr){2-11}

    & \multirow{2}{*}{CIFAR-10~(EDM)} & Baseline & \B 2.74 & \B 0.604 & \B 3.21 & 0.597 & 4.48 & 0.586 & 9.51 & 0.531 \\
    & & OGDM (ft) & 3.75 & \B 0.604 & 3.62 & \B 0.605 & \B 3.60 & \B 0.600 & \B 4.97 & \B 0.586 \\
    \cmidrule(lr){2-11}

    & \multirow{2}{*}{CIFAR-10~(EDM cond.)} 
    & Baseline & \B 2.50 & 0.606 & 2.87 & 0.599 & 3.76 & 0.584 & 6.63 & 0.544 \\
    & & OGDM (ft) & 2.87 & \B 0.610 & \B 2.75 & \B 0.607 & \B 2.69 & \B 0.601 & \B 3.53 & \B 0.573 \\
    \cmidrule(lr){2-11}
    
    & \multirow{2}{*}{CelebA~(ADM)} & Baseline & 3.67 & 0.553 & 4.15 & 0.539 & 5.22 & 0.511 & 7.33 & 0.445 \\
    &  & OGDM & \B 2.62 & \B 0.607 & \B 2.70 & \B 0.604 & \B 2.96 & \B 0.585 & \B 4.35 & \B 0.545 \\
    \cmidrule(lr){2-11}

    & \multirow{2}{*}{LSUN Church~(LDM)} 
    & Baseline & 8.41 & 0.470 & 8.21 & 0.471 & 8.07 & 0.475 & 9.14 & 0.464 \\
    & & OGDM (ft) & 
    \B 7.69 & \B 0.480 & \B 7.48 & \B 0.489 & \B 7.48 & \B 0.481 & \B 8.68 & \B 0.478 \\
    \hline\hline

    \multirow{7}{*}{F-PNDM} & \multirow{2}{*}{CIFAR-10~(ADM)} 
    & Baseline & \B 5.17 & 0.609 & \B 6.19 & 0.600 & 10.55 & 0.535 & -- & -- \\
    & & OGDM & 5.40 & \B 0.609 & 6.72 & \B 0.600 & \B 8.84 & \B 0.557 & -- & -- \\
    \cmidrule(lr){2-11}
    
    & \multirow{2}{*}{CelebA~(ADM)} & Baseline & 3.39 & 0.562 & 4.25 & 0.539 & 7.08 & 0.488 & -- & -- \\
    &  & OGDM & \B 2.81 & \B 0.615 & \B 3.10 & \B 0.609 & \B 5.14 & \B 0.576 & -- & -- \\
    \cmidrule(lr){2-11}

    & \multirow{2}{*}{LSUN Church~(LDM)} & Baseline & 9.04 & 0.474 & 9.10 & 0.483 & 12.75 & 0.493 & -- & -- \\
    & & OGDM (ft) & 
    \B 8.24 & \B 0.481 & \B 8.39 & \B 0.495 & \B 11.78 & \B 0.505 & -- & --  \\
    \bottomrule    
        \end{tabular}
    }
\vspace{3mm}
\end{table*}
\begin{table*}[t]
    \centering
    \caption{
        Comparisons of FIDs with other methods on CIFAR-10 (32$\times$32) and CelebA (64$\times$64).
        `$\dagger$' means the values are copied from other papers and `*' means the values are obtained by applying DDIM~\cite{song2021denoising} as a sampler.
    }
    \label{tab:comparison}
    \vspace{-1mm}
    \setlength{\tabcolsep}{3.6mm}
    \scalebox{0.90}{
        \begin{tabular}{lccccccccc}
            \toprule
            & \multicolumn{4}{c}{CIFAR-10} & ~ & \multicolumn{4}{c}{CelebA}\\  
            \cmidrule(lr){2-5} \cmidrule(lr){7-10}   
            Method ~\textbackslash~ NFEs & 25 & 20 & 15 & 10 & ~ & 25 & 20 & 15 & 10 \\ \hline

            DDIM~\cite{song2021denoising} $\dagger$* & -- & 6.84 & -- & 13.36 & ~ & -- & 13.73 & -- & 17.33 \\
            Analytic-DPM~\cite{bao2022analytic} $\dagger$* & 5.81 & -- & -- & 14.00 & ~ & 9.22 & -- & -- & 15.62 \\
            FastDPM~\cite{kong2021on} $\dagger$* & -- & 5.05 & -- & 9.90 & ~ & -- & 10.69 & -- & 15.31 \\
            GENIE~\cite{dockhorn2022genie} $\dagger$ & 3.64 & 3.94 & 4.49 & 5.28 & ~ & -- & -- & -- & -- \\
            Watson \etal~\cite{watson2022learning} $\dagger$ & 4.25 & 4.72 & 5.90 & 7.86 & ~ & -- & -- & -- & -- \\

            ADM~\cite{dhariwal2021diffusion} & 7.08 & 8.05 & 9.93 & 15.20 & ~ & 7.20 & 7.88 & 9.34 & 11.92 \\
            EDM~\cite{karras2022elucidating} & 2.19 & -- & 4.48 & -- & ~ & -- & -- & -- & -- \\

            S-PNDM~\cite{liu2022pseudo} & 2.74 & \B 3.21 & 4.48 & 9.51 & ~ & 3.67 & 4.15 & 5.22 & 7.33 \\
            F-PNDM~\cite{liu2022pseudo} & 5.17 & 6.19 & 10.55 & -- & ~ & 3.39 & 4.25 & 7.08 & -- \\
            CT~\cite{song2023consistency} & 6.94 & 6.63 & 6.36 & 6.20 & ~ & -- & -- & -- & -- \\

            \hline 
            OGDM & \B 2.17 & 3.53 & \B 3.60 & \B 4.97 & ~ & \B 2.62 & \B 2.70 & \B 2.96 & \B 4.35 \\
            \bottomrule    
        \end{tabular}
    }
\end{table*}

\subsection{Quantitative results}
\label{subsec:quantitative}

\cref{tab:quant_euler,tab:quant_heun} demonstrate quantitative comparison results when projection functions $f_\theta(\cdot)$ aligns with samplers well. 
They show that a proper combination of a projection function and a sampler substantially improves FID and recall in all cases with NFEs~$\leq25$.
This observation implies that the proposed method yields a robust denoising network for large step sizes. 
\cref{tab:quant_euler} also compares performance between our models trained from scratch and the ones fine-tuned on `CIFAR-10 (ADM)' and `CelebA (ADM)'. 
We observe that the fine-tuned models exhibit competitive performance with a small fraction (5--10\%) of the training iterations when compared to the models optimized through full training. 
Detailed analysis and comparisons are provided in \cref{tab:hyperparameter2} of \cref{app:hyperparameter}.
These findings highlight the practicality and computational efficiency of the proposed method.

\cref{tab:quant_pndm} presents FID and recall scores in the case that the projection function $f_\theta(\cdot)$ is a step of the Euler method while the samplers are either S-PNDM or F-PNDM.
Unless the projection function and the sampler align properly, the benefit from our approach is not guaranteed because the observations may be inaccurately projected onto the manifold from the perspective of the sampler.
Despite this reasonable concern, both S-PNDM and F-PNDM still achieve great performance gains especially when NFEs are small.
Notably, the combination of the proposed method and S-PNDM shows consistent performance improvements; the models with the Euler projection and S-PNDM harness synergy because the steps of S-PNDM are similar to the Euler method except for the initial step.
Moreover, `CIFAR-10~(EDM cond.)' in \cref{tab:quant_euler,tab:quant_pndm} shows that our method is still effective for conditional sampling.
We also present results when stochastic sampling is employed, with our approach showing superior performance, in \cref{app:stochastic}.

\cref{tab:comparison} presents the FIDs of various algorithms combined with fast inference techniques in a wide range of NFEs.
The results imply that the proposed approach is advantageous when the number of time steps for inference is small.
We additionally compare our approach with other methods that incorporate time-dependent discriminators to diffusion process, \ie, DG~\cite{kim2023refining} and DDGAN~\cite{xiao2022DDGAN}, in Section~\cref{app:compare_gan}.

\subsection{Qualitative results}
\label{subsec:qualitative}

We discuss the qualitative results of our approach in the following two aspects.

\vspace{-2mm}
\paragraph{Comparisons to baselines}
We provide qualitative results on CIFAR-10, CelebA, and LSUN Church obtained by a few sampling steps in comparison with the baseline methods in \cref{app:qual}.
The baseline models often produce blurry samples when utilizing fast inference methods. 
In contrast, our models generate crispy and clear images as well as show more diverse colors and tones compared to the corresponding baselines.
Moreover, as shown in the left-hand side of \cref{fig:teaser}, the baseline model generates face images with inconsistent genders by varying NFEs.
On the other hand, our model maintains the information accurately, which is desirable results because we use deterministic generative process with the same initial point.
This is because the additional loss term of our method enables the model to approximate each backward step more accurately, even with coarse discretization. 

\vspace{-2mm}
\paragraph{Nearest neighborhoods}
\cref{fig:nn} in \cref{app:nn} illustrates the nearest neighbor examples in the training datasets with respect to the generated images by our approach.
According to the results, the generated samples are sufficiently different from the training examples, confirming that our models do not simply memorize data but increase scores properly by improving  diversity in output images.


\vspace{3mm}
\subsection{Discussion on training cost}
\label{sub:discussion}

Regarding the trainig cost, our method increases the training time by approximately 80\% compared to the baseline, given the same number of iterations.
This is primarily due to the additional training cost incurred by the discriminator. 
However, we can achieve promising results by fine-tuning the pretrained baseline model for only a small number of iterations, as shown in \cref{tab:quant_euler,tab:quant_heun,tab:quant_pndm}, which significantly alleviate the burden of training the discriminator.

\section{Future work}
\label{sec:future}

Although not explored in this work, there is more room for amplifying the impact of the proposed approach.
Using OGDM as a pretrained score model for consistency distillation~\cite{song2023consistency} can be advantageous over using baseline models by enhancing the accuracy of one-step progress in the teacher model.
Moreover, integrating the lookahead variable $s$ as an additional input to diffusion models may further improve performance.
We can also construct a specialized denoising network for specific sampling steps by focusing more on learning manifolds of noise levels corresponding to the time steps to be used in sampling.
This can be realized by selecting $s$ adaptively, rather than adopting a uniform sampling.
Moreover, considering that $\xi(\beta)$ is positively correlated with $\beta$ (see \cref{fig:proposition1}~(b) in \cref{app:proposition1} of the supplementary document), it would make sense to set $\gamma$ in proportion to $1-\frac{\bar\alpha_t}{\bar\alpha_{t-s}}$.
While we examine the observations following the Bernoulli distribution and implement them using the discriminators as in GANs, it is important to note that there are other options to define and implement the observation factors.
For example, other formulations of the extended surrogate~\cref{eq:intermediate objective} can be explored, which may include the ones specific to target tasks, available resources, and measurement methods.

\section{Conclusion}
\label{sec:conclusion}

We presented a diffusion probabilistic model that introduces observations into the Markov chain of \cite{ho2020denoising}.
As a feasible and effective way, we have concretized the surrogate loss for negative log-likelihood using observations following the Bernoulli distribution, and integrated the adversarial training loss by adding a discriminator network that simulates the observation. 
Our strategy regulates the denoising network to minimize the accurate negative log-likelihood surrogate at inference, thereby increasing robustness in a few steps sampling.
As a result, our method facilitates faster inference by mitigating quality degradation.
We demonstrated the effectiveness of the proposed method on well-known baseline models and multiple datasets.

\paragraph{Acknowledgements}
This work was partly supported by LG AI Research, and the IITP grants [No. 2022-0-00959, (Part 2) Few-Shot Learning of Causal Inference in Vision and Language for Decision Making; 
No. 2021-0-02068, AI Innovation Hub (AI Institute, Seoul National University); No. 2021-0-01343, Artificial Intelligence Graduate School Program (Seoul National University)] funded by the Korea government (MSIT).

{
    \small
    \bibliographystyle{ieeenat_fullname}
    \bibliography{main}
}


\clearpage
\onecolumn
\begin{center}
    \Large {\textbf{Observation-Guided Diffusion Probabilistic Models} \\ 
                    \textit{Supplementary Material}} \\
\end{center}
\appendix


\section*{\Large Appendix}
\section{Details of \cref{sec:formulation}}
\subsection{Proof of \cref{eq:forward}}
\begin{align*}
	q(\x_{1:T}, \y_{0:T} | \x_0) 
    &= q(\y_0|\x_0) \prod_{t=0}^{T-1} q(\x_{t+1}|\x_{0:t},\y_{0:t}) q(\y_{t+1}|\x_{0:t+1},\y_{0:t})\\  
    &= q(\y_0|\x_0) \prod_{t=0}^{T-1} q(\x_{t+1}|\x_{t}) q(\y_{t+1}|\x_{t+1}) \qquad (\because \cref{eq:prop_forward})\\
    &= \prod_{t=0}^T q(\y_{t} | \x_{t}) \prod_{t=0}^{T-1} q(\x_{t+1}|\x_{t}) \\
    &= \prod_{t=0}^T q(\y_{t} | \x_{t}) \left[q(\x_1|\x_0)\prod_{t=1}^{T-1} q(\x_{t+1}|\x_{t}, \x_{0})\right] \qquad (\because \cref{eq:prop_forward})\\
    &= \prod_{t=0}^T q(\y_{t} | \x_{t}) \left[q(\x_1|\x_0)\prod_{t=1}^{T-1} \frac{q(\x_{t+1},\x_{t}| \x_{0})}{q(\x_{t}| \x_{0})}\right] \\
    &= \prod_{t=0}^T q(\y_{t} | \x_{t}) \left[q(\x_1|\x_0)\prod_{t=1}^{T-1} \frac{q(\x_{t+1}|\x_{0})q(\x_{t}| \x_{t+1},\x_{0})}{q(\x_{t}| \x_{0})}\right] \\
	&= q(\x_T | \x_0) \prod_{t=2}^T q(\x_{t-1} | \x_t, \x_0) \prod_{t=0}^T q(\y_{t} | \x_{t}). 
\end{align*}
\subsection{Proof of \cref{eq:backward}}
\begin{align*}
    p(\x_{T:0},\y_{T:0}) 
    &= p(\x_T)p(\y_{T}|\x_{T}) \prod_{t=T}^{1} p(\x_{t-1}|\x_{T:t},\y_{T:t})p(\y_{t-1}|\x_{T:t-1},\y_{T:t}) \\
    &= p(\x_T)p(\y_{T}|\x_{T}) \prod_{t=T}^{1} p(\x_{t-1}|\x_t)p(\y_{t-1}|\x_{t-1}) \qquad (\because \cref{eq:prop_forward}) \\
    &= p(\x_T) \prod_{t=T}^{1} p(\x_{t-1}|\x_t) \prod_{t=T}^{0} p(\y_{t} | \x_{t}).
\end{align*}
\subsection{Proof of \cref{lemma1}}
\label{app:lemma1}
\limit*
\begin{proof}[Proof of \cref{eq:asymptotic1}]

Let $\Vu' = \sqrt{1-\beta}\Vu$. Then,
\begin{align*}
    p_{\Vv|\Vu'}(\Vv|\Vu') 
    = \mathcal{N}(\Vv;\Vu', \beta\mathbf{I}) 
    = (2\pi\beta)^{-d/2}\exp(-\frac{1}{2\beta}||\Vv-\Vu'||^2) 
    = \mathcal{N}(\Vu';\Vv, \beta\mathbf{I}) 
    = q_{\Vu'|\Vv}(\Vu'|\Vv).
\end{align*}
By talyor expansion of $p_{\Vu'}(\Vu')$ at $\Vv$,
\begin{align*}
    p_{\Vu'}(\Vu') = p_{\Vu'}(\Vv) + <\nabla p_{\Vu'}(\Vv), \Vu' - \Vv> + O(||\Vu'-\Vv||^2).
\end{align*}
Then,
\begin{align*}
    \int q_{\Vu'|\Vv}(\Vu'|\Vv)p_{\Vu'}(\Vu') d\Vu' 
    &= \mathbb{E}_{\Vu' \sim \mathcal{N}(\Vv, \beta\mathbf{I})}[p_{\Vu'}(\Vv) + <\nabla p_{\Vu'}(\Vv), \Vu' - \Vv> + O(||\Vu'-\Vv||^2)] \\
    &= p_{\Vu'}(\Vv) + O(\beta).
\end{align*}
By Bayes' rule and above result,
\begin{align*}
    p_{\Vu'|\Vv}(\Vu'|\Vv) 
    &= \cfrac{p_{\Vv|\Vu'}(\Vv|\Vu')p_{\Vu'}(\Vu')}{\int p_{\Vv|\Vu'}(\Vv|\Vu')p_{\Vu'}(\Vu') d\Vu'} \\
    &= \cfrac{q_{\Vu'|\Vv}(\Vu'|\Vv)p_{\Vu'}(\Vu')}{\int q_{\Vu'|\Vv}(\Vu'|\Vv)p_{\Vu'}(\Vu') d\Vu'} \\
    &= q_{\Vu'|\Vv}(\Vu'|\Vv) \cfrac{p_{\Vu'}(\Vv) + <\nabla p_{\Vu'}(\Vv), \Vu' - \Vv> + O(||\Vu'-\Vv||^2)}{p_{\Vu'}(\Vv) + O(\beta)} \\
    &= q_{\Vu'|\Vv}(\Vu'|\Vv) (1+ <\frac{\nabla p_{\Vu'}(\Vv)}{p_{\Vu'}(\Vv)}, \Vu' - \Vv> + O(||\Vu'-\Vv||^2))(1 + O(\beta)) \\
    &= q_{\Vu'|\Vv}(\Vu'|\Vv) \exp(<\nabla \log p_{\Vu'}(\Vv), \Vu' - \Vv>) + O(\beta) \\
    &= (2\pi\beta)^{-d/2}\exp(-\frac{1}{2\beta}||\Vv-\Vu'||^2) \exp(<\nabla \log p_{\Vu'}(\Vv), \Vu' - \Vv>) + O(\beta) \\
    &= (2\pi\beta)^{-d/2}\exp(-\frac{1}{2\beta}(||\Vv-\Vu'||^2 -2\beta<\nabla \log p_{\Vu'}(\Vv), \Vu' - \Vv>)) + O(\beta) \\
    &= (2\pi\beta)^{-d/2}\exp(-\frac{1}{2\beta}||\Vu' - \Vv -\beta \nabla\log p_{\Vu'}(\Vv)||^2 + O(\beta)) + O(\beta) \\
    &\approx \mathcal{N}(\Vu'; \Vv + \beta \nabla \log p_{\Vu'}(\Vv),\beta \mathbf{I}) ~\text{for}~ \beta \ll 1.
\end{align*}
From $\Vu' = \sqrt{1-\beta}\Vu$,
\begin{align*}
    p_{\Vu|\Vv}(\Vu|\Vv) 
    &= \mathcal{N}(\Vu; \frac{1}{\sqrt{1-\beta}} (\Vv + \beta \nabla \log p_{\Vu}(\Vv)),\frac{\beta}{1-\beta} \mathbf{I}) \\
    &\approx \mathcal{N}(\Vu; \frac{1}{\sqrt{1-\beta}} (\Vv + \beta \nabla \log p_{\Vu}(\Vv)),\beta \mathbf{I}) ~\text{for}~ \beta \ll 1\qedhere
\end{align*}
\end{proof}
\begin{proof}[Proof of \cref{eq:asymptotic2}]
\begin{align*}
    \lim_{\beta \rightarrow 1^-} p_{\Vv|\Vu}(\Vv|\Vu) 
    &= \lim_{\beta \rightarrow 1^-} (2\pi\beta)^{-d/2} \exp(-\frac{1}{2\beta}||\Vv-\sqrt{1-\beta}\Vu||^2) \\
    &= (2\pi)^{-d/2} \exp(-\frac{1}{2}||\Vv||^2) \coloneqq f(\Vv). \\
\end{align*}
From Bayes' rule and above results,
\begin{align*}
    \therefore \lim_{\beta \rightarrow 1^-} p_{\Vu|\Vv}(\Vu|\Vv) 
    &= \lim_{\beta \rightarrow 1^-} \cfrac{p_{\Vv|\Vu}(\Vv|\Vu)p_\Vu(\Vu)}{\int p_{\Vv|\Vu}(\Vv|\Vu)p_\Vu(\Vu) d\Vu} \\
    &= \cfrac{ \lim_{\beta \rightarrow 1^-} p_{\Vv|\Vu}(\Vv|\Vu)p_\Vu(\Vu)}{\int \lim_{\beta \rightarrow 1^-} p_{\Vv|\Vu}(\Vv|\Vu)p_\Vu(\Vu) d\Vu} \\
    &= \cfrac{f(\Vv)p_\Vu(\Vu)}{\int f(\Vv)p_\Vu(\Vu) d\Vu} \\
    &= \cfrac{p_\Vu(\Vu)}{\int p_\Vu(\Vu) d\Vu} = p_\Vu(\Vu)\qedhere
\end{align*}
\end{proof}

\subsection{Behaviors of $p_{\bf{u}|\bf{v}}^{(\beta)}(\bf{u}|\bf{v})$, $q_{\bf{u}|\bf{v}}^{(\xi)}(\bf{u}|\bf{v})$, and $\xi(\beta)$}
\label{app:proposition1}
In this section, we justify the approximation on the density function of reverse distribution by showing behaviors of $p_{\Vu|\Vv}^{(\beta)}(\Vu|\Vv)$, $q_{\Vu|\Vv}^{(\xi)}(\Vu|\Vv)$, and $\xi(\beta)$ on toy examples.

Followings are the definitions in \cref{subsec:analysis_final_obj}.
For $\Vu \sim p_\Vu$ and $\Vv|\Vu \sim \mathcal{N}(\sqrt{1-\beta}\Vu, \beta \mathbf{I})$, $p_{\Vu|\Vv}^{(\beta)}(\Vu|\Vv)$ is a real backward density function,
\begin{align*}
	q_{\Vu|\Vv}^{(\xi)}(\Vu|\Vv) = C (\mathcal{N}(\Vu;\frac{1}{\sqrt{1-\beta}}(\Vv + \beta \nabla\log p_\Vu(\Vv)), \beta \mathbf{I}))^{1-\xi} p(\Vu)^{\xi},
\end{align*}
and 
\begin{align*}
	\xi(\beta) \in \argmin_{\xi \in [0,1]} \int_{-\infty}^{\infty} (q_{\Vu|\Vv}^{(\xi)}(\Vu|\Vv) - p_{\Vu|\Vv}^{(\beta)}(\Vu|\Vv))^2 d\Vu.
\end{align*}

For arbitrary $\mu>0$, let
\begin{align*} 
    p_{\Vu}(\Vu) 
    &= \frac{1}{2} (\mathcal{N}(\Vu;\mu,1) + \mathcal{N}(\Vu;-\mu,1)) \\
    &= \frac{1}{2} (2\pi)^{-1/2} (\exp(-(\Vu-\mu)^2/2 ) + \exp(-(\Vu+\mu)^2/2 )).
\end{align*}
Then, $\mathcal{N}(\Vu;\frac{1}{\sqrt{1-\beta}}(\Vv + \beta \nabla\log p_\Vu(\Vv)), \beta)$, and $p_{\Vu|\Vv}^{(\beta)}(\Vu|\Vv) = \cfrac{p_{\Vv|\Vu}(\Vv|\Vu)p_\Vu(\Vu)}{\int p_{\Vv|\Vu}(\Vv|\Vu)p_\Vu(\Vu) d\Vu}$ can be explicitly expressed.
Moreover, $q_{\Vu|\Vv}^{(\xi)}(\Vu|\Vv)$ can be calculated numerically. 
Therefore, we can numerically calculate $\ell^2$-norm between $p_{\Vu|\Vv}^{(\beta)}(\Vu|\Vv)$ and $q_{\Vu|\Vv}^{(\xi)}(\Vu|\Vv)$.
Finally, we can obtain $\xi(\beta)$ for each $\Vv$ and $\beta$.
\begin{figure*}[th!]
    \centering
    \setlength{\tabcolsep}{0.2mm}
    \scalebox{0.72}{
    \begin{tabular}{ccc}
        \includegraphics[width=0.46\textwidth]{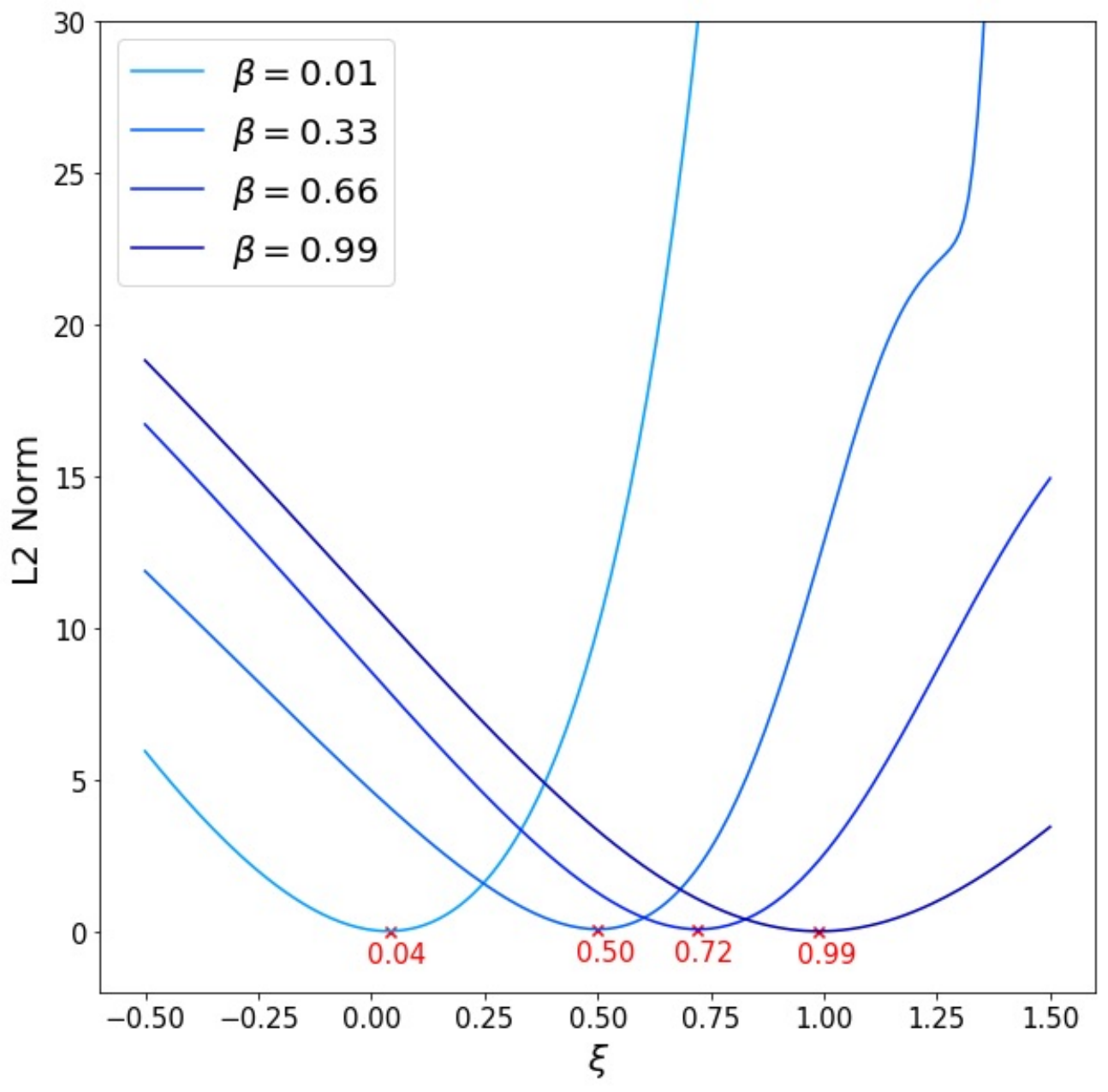} &
        \includegraphics[width=0.46\textwidth]{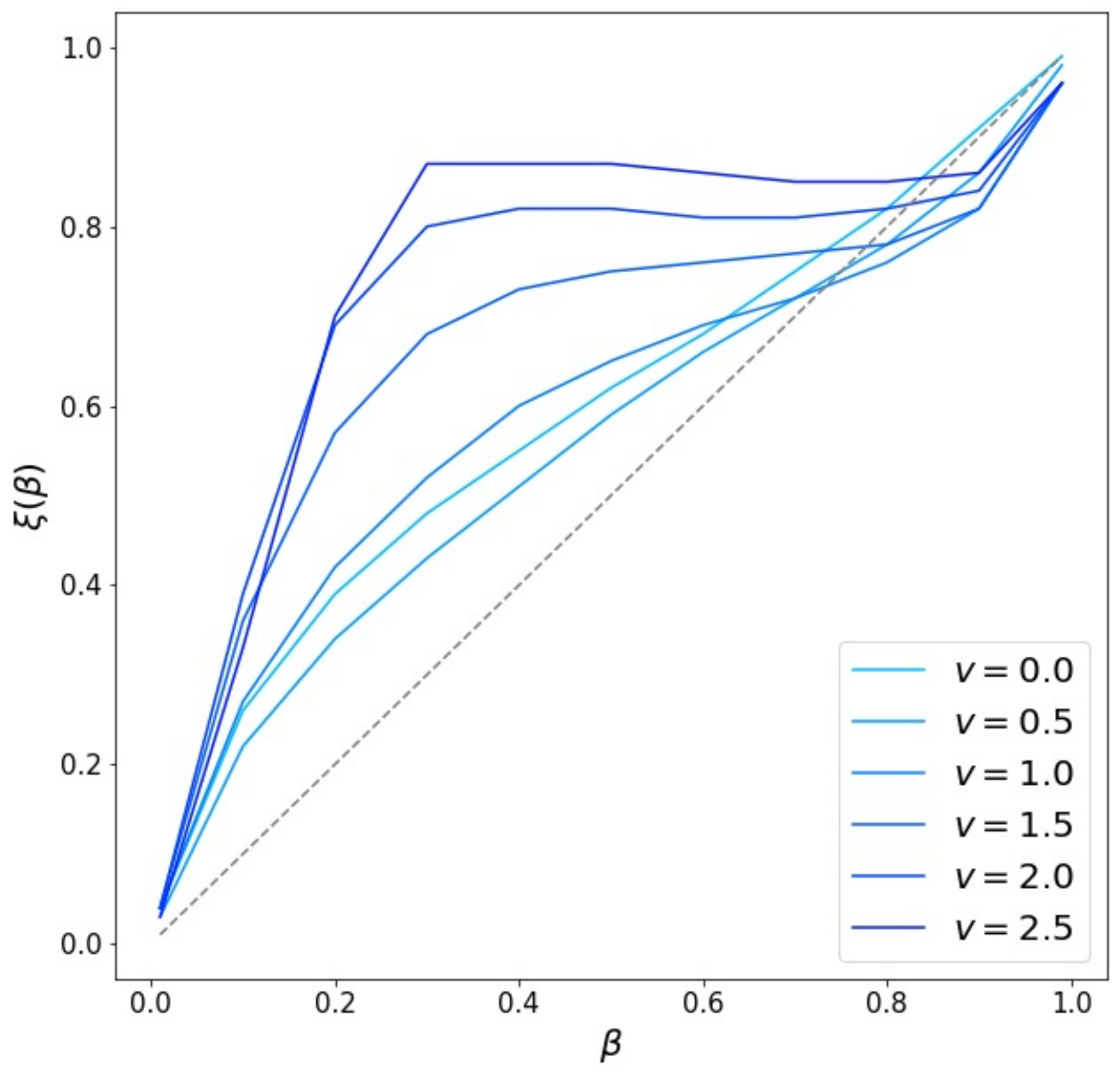} &
        \includegraphics[width=0.46\textwidth]{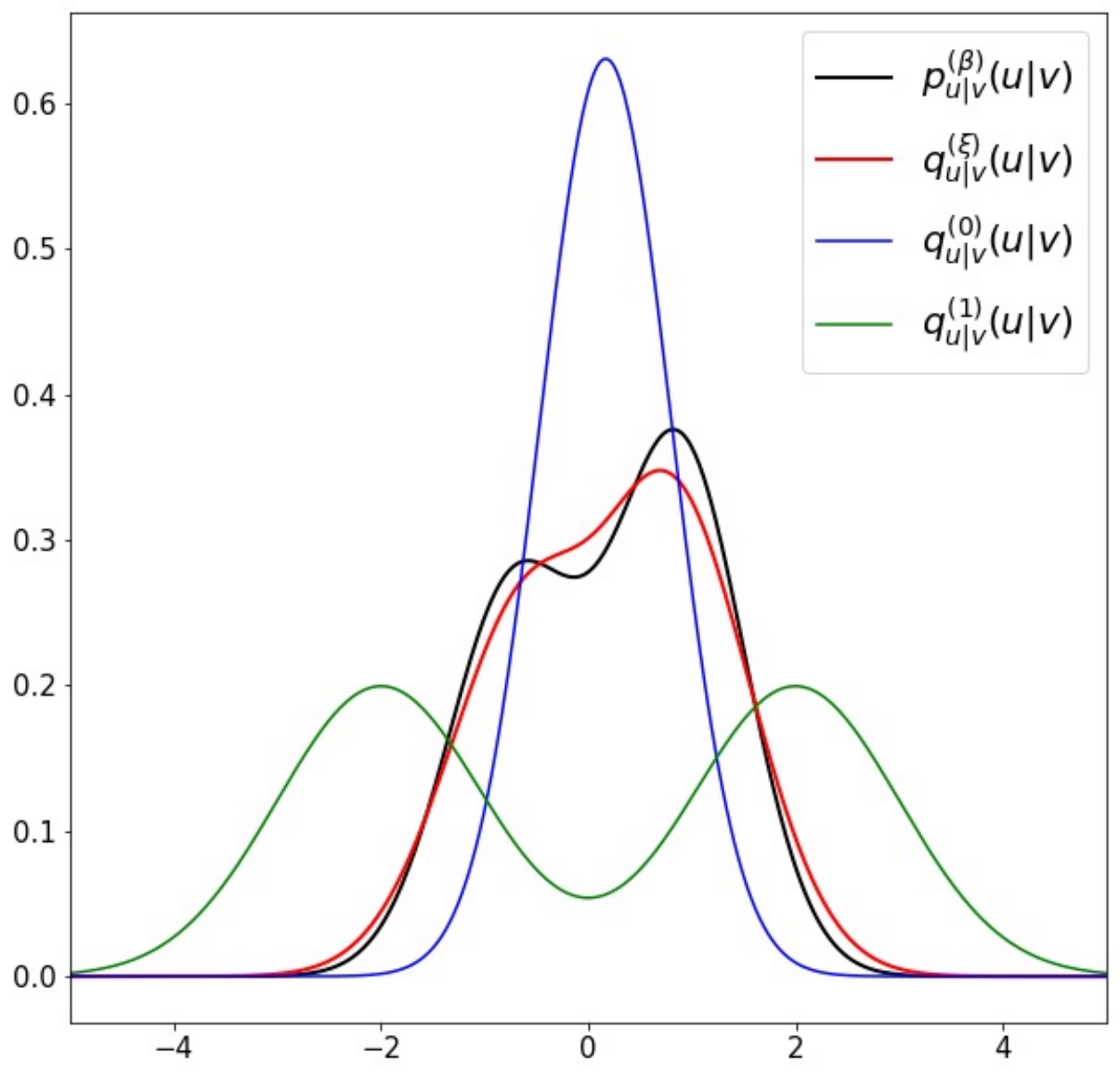} \\
        (a) & (b) & (c) 
    \end{tabular}
    }
    \caption{
        Simulations when $\mu=2$.
        (a) $\ell^2$-norm between $p_{\Vu|\Vv}^{(\beta)}(\Vu|\Vv)$, and $q_{\Vu|\Vv}^{(\xi)}(\Vu|\Vv)$ with respect to $\xi$ when $\Vv=0.1$. 
        (b) The graph of $\xi(\beta)$ with respect to $\beta$ for various $\Vv$. 
        (c) Pdfs of four distributions when $\Vv=0.1$, $\beta=0.4$, and $\xi=\xi(\beta)=0.55$. 
	}
    \label{fig:proposition1}  
\end{figure*}

\cref{fig:proposition1}(a) shows the that $q_{\Vu|\Vv}^{(\xi(\beta))}(\Vu|\Vv)$ better approximates $p_{\Vu|\Vv}^{(\beta)}(\Vu|\Vv)$ than $q_{\Vu|\Vv}^{(1)}(\Vu|\Vv)=p_{\Vu}(\Vu)$ and $q_{\Vu|\Vv}^{(0)}(\Vu|\Vv)=\mathcal{N}(\Vu;\frac{1}{\sqrt{1-\beta}}(\Vv + \beta \nabla\log p_\Vu(\Vv)), \beta)$.
We can observe that $0 < \xi(\beta) < 1$ for $\forall \beta \in (0,1)$ from \cref{fig:proposition1}(b).
In \cref{fig:proposition1}(c), the black line is the precise reverse distribution while the blue line is asymptotic function when $\beta \rightarrow 0^+$ and the green line is the limit when $\beta \rightarrow 1^-$.
Note that the blue line is the approximation used in vanilla diffusion models.
The red line which is a normalized weighted geometric mean of blue and green lines better approximates the real distribution. 

\clearpage
\section{Numerical ODE solvers}
\label{app:solvers}
Given discretization steps of $T=t_N > t_{k-1} > \cdots > t_0 = 0$, \cref{alg:euler_method,alg:heun_method} caculate the numerical solution at $t=0$ with initial condition $\x_T$ for the following ODE:  
\begin{align}
    d\x_t = f(\x_t, t)dt.
\end{align}

\RestyleAlgo{ruled}
\begin{algorithm}
    \caption{Euler Method}\label{alg:euler_method}
    \For{$i=N, \cdots ,1$}{
        $\x_{t_{i-1}} = \x_{t_{i}} + (t_{i-1} - t_{i})f(\x_{t_i},t_i)$
    }
    \Return{$\x_0$}
\end{algorithm}
\vspace{-7mm}
\begin{algorithm}
    \caption{Heun's Method}\label{alg:heun_method}
    \For{$i=N, \cdots ,1$}{
        $\x_{t_{i-1}} = \x_{t_{i}} + (t_{i-1} - t_{i})f(\x_{t_i},t_i)$ \\
        \If{$i > 1$}{
            $\x_{t_{i-1}} = \x_{t_{i}} + (t_{i-1} - t_{i}) (\frac{1}{2}f(\x_{t_i},t_i) + \frac{1}{2}f(\x_{t_{i-1}},t_{i-1}))$
        }
    }
    \Return{$\x_0$}
\end{algorithm}
\vspace{-2mm}
\section{Hyperparameters for experiments}
\label{app:hyperparameter}
\vspace{-3mm}
\begin{table*}[h!]
    \centering
    \caption{Hyperparameters for Training}
    \label{tab:hyperparameter}
        \setlength{\tabcolsep}{2mm}
    \scalebox{0.85}{
        \begin{tabular}{cclccccc}
            \toprule
            Dataset & Backbone & Training & Projection &$k$ & $\gamma$ & Batch size & Seen Images \\
            \hline

            \multirow{7}{*}{CIFAR-10} & \multirow{3}{*}{ADM} & Baseline & -- & -- & -- & 128 & 38M \\
            & & OGDM & Euler & 0.1 & 0.01 & 128 & 38M \\
            & & OGDM (ft) & Euler & 0.2 & 0.025 & 128 & 2M \\
            \cmidrule(lr){2-8}
            
            & \multirow{3}{*}{EDM} & Baseline & -- & -- & -- & 512 & 200M \\
            & & OGDM (ft) & Euler & 0.2 & 0.025 & 512 & 20M \\
            & & OGDM (ft) & Heun's & 0.2 & 0.005 & 512 & 20M \\
            \hline
            
            \multirow{3}{*}{CelebA} & \multirow{3}{*}{ADM} & Baseline & -- & -- & -- & 128 & 38M \\
            & & OGDM & Euler & 0.1 & 0.01 & 128 & 38M \\
            & & OGDM (ft) & Euler & 0.2 & 0.025 & 128 & 2M \\
            \hline 
            \multirow{2}{*}{LSUN Church} & \multirow{2}{*}{LDM} & Baseline & -- & -- & -- & 96 & 48M \\
            & & OGDM (ft) & Euler & 0.1 & 0.01 & 96 & 1.5M\\
            \bottomrule
        \end{tabular}
        }
\end{table*}
\vspace{-4mm}
\begin{table*}[h!]
    \centering
    \caption{Hyperparameters for sampling}
    \label{tab:hyperparameter2}
        \setlength{\tabcolsep}{2mm}
    \scalebox{0.85}{
        \begin{tabular}{cccc}
            \toprule
            Dataset & Backbone & Sampler & Discretization \\
            \hline

            \multirow{7}{*}{CIFAR-10} & \multirow{3}{*}{ADM} & Euler & quadratic\\
            & & S-PNDM & linear \\
            & & F-PNDM & linear \\
            \cmidrule(lr){2-4}
            
            & \multirow{3}{*}{EDM} & Euler & default of \cite{karras2022elucidating} \\
            & & Heun's & default of \cite{karras2022elucidating} \\
            & & S-PNDM & default of \cite{karras2022elucidating} \\
            \hline
            
            \multirow{3}{*}{CelebA} & \multirow{3}{*}{ADM} & Euler & linear \\
            & & S-PNDM & linear \\
            & & F-PNDM & linear \\
            \hline 
             
            \multirow{3}{*}{LSUN Church} & \multirow{3}{*}{LDM} & Euler & linear \\
            & & S-PNDM & quadratic \\
            & & F-PNDM & quadratic \\
            \bottomrule
        \end{tabular}
        }
\end{table*}

\clearpage 
\section{Comparisons with DG and DDGAN}
\label{app:compare_gan}
In this section, we compare OGDM with DG~\cite{kim2023refining} and DDGAN~\cite{xiao2022DDGAN}; they adopt time-dependent discriminators in either training or inference. 

\subsection{Comparison with DG}
\label{app:compare_dg}
\begin{table*}[h]
    \centering
    \caption{
    FID and recall scores of two samplers: Heun's method, DG sampler. 
    Two samplers are applied to vanilla diffusion models and OGDM for variuos discretization steps.
    For each step, diffusion models are evaluated twice and the gradient of a discriminator is calculated once for DG sampler. 
    The time to calculate the gradient of the discriminator requires about 180\% time of evaluating the diffusion models, but it is regarded as $\text{NFEs}=1$ in the table. 
    The number of steps~($n$) are chosen by $\argmin_{n} 3n-2 \geq (35, 25, 20, 15)$.
    }
    \label{tab:dg}
    \vspace{-2mm}
        \setlength{\tabcolsep}{2.4mm}
    \scalebox{0.85}{
        \begin{tabular}{clccccccccc}
    \toprule
    \multicolumn{2}{c}{\diagbox[height=1.2\line]{}{\hspace{15mm} \# steps~($n$)}} & 
    \multicolumn{2}{c}{13} & \multicolumn{2}{c}{9} & \multicolumn{2}{c}{8} & \multicolumn{2}{c}{6} &  \\ 
    \cmidrule(lr){3-4}\cmidrule(lr){5-6}\cmidrule(lr){7-8}\cmidrule(lr){9-10}  
    Dataset~(Backbone) & Method & FID$\downarrow$ & Rec.$\uparrow$ & FID$\downarrow$ & Rec.$\uparrow$ & FID$\downarrow$ & Rec.$\uparrow$ & FID$\downarrow$ & Rec.$\uparrow$ & NFEs \\ \hline
    
    \multirow{4}{*}{CIFAR-10~(EDM)}
    & Baseline & 2.19 & 0.616 & 3.33 & 0.615 & 4.48 & 0.604 & 15.69 & 0.535 & $2n-1$  \\
    & Baseline + DG~\cite{kim2023refining} & \B 1.99 & 0.630 & 4.62 & 0.613 & 7.39 & 0.586 & 24.78 & 0.465 & $3n-2$ \\
    & OGDM (ft) & 2.17 & 0.622 & \B 2.99 & 0.622 & \B 4.21 & \B 0.619 & \B 13.59 & \B 0.591 & $2n-1$ \\
    & OGDM (ft) + DG~\cite{kim2023refining} & 2.00 & \B 0.633 & 3.58 & \B 0.624 & 5.77 & 0.616 & 19.25 & 0.560 & $3n-2$\\

    \bottomrule    
        \end{tabular}
        }
\end{table*}
DG utilizes discriminator to improve the sample quality without taking sampling efficiency into account.
\cref{tab:dg} demonstrates quantitative comparison results between DG and OGDM.
DG does not work well for fast sampling, rather it deteriorates the sample quality of a few-step-sampler.
Moreover, DG requires more computational cost for each update since it incorporates gradient calculation.
\subsection{Comparison with DDGAN}
\label{app:compare_ddgan}

\begin{table*}[h]
    \centering
    \caption{
        FID and recall scores of DDGAN~\cite{xiao2022DDGAN} and OGDM on CIFAR-10. `$\dagger$' means the values are copied from the literature.
        }
    \label{tab:ddgan}
    \vspace{-2mm}
        \setlength{\tabcolsep}{2.4mm}
    \scalebox{0.85}{
        \begin{tabular}{lcccccccccc}
    \toprule
    \diagbox[height=1.2\line]{}{NFEs}& 
    \multicolumn{2}{c}{20} & \multicolumn{2}{c}{8} & \multicolumn{2}{c}{4} & \multicolumn{2}{c}{2} & \multicolumn{2}{c}{1} \\ 
    \cmidrule(lr){2-3}\cmidrule(lr){4-5}\cmidrule(lr){6-7}\cmidrule(lr){8-9}\cmidrule(lr){10-11}  
    Method & FID$\downarrow$ & Rec.$\uparrow$ & FID$\downarrow$ & Rec.$\uparrow$ & FID$\downarrow$ & Rec.$\uparrow$ & FID$\downarrow$ & Rec.$\uparrow$ & FID$\downarrow$ & Rec.$\uparrow$ \\ \hline
    
    DDGAN~\cite{xiao2022DDGAN}$\dagger$ & -- & -- & \B 4.36 & 0.56 & 3.75 & 0.57 & 4.08 & 0.54 & 14.60 & 0.19\\
    OGDM & 3.53 & 0.60 & 6.16 & \B 0.58 & -- & -- & -- & -- & -- & -- \\

    \bottomrule    
        \end{tabular}
        }
\end{table*}
DDGAN parametrizes the reverse distribution by neural network without Gaussian assumption.
Its objective is not log-likelihood driven and therefore it is more a variant of GANs rather than a diffusion model.
\cref{tab:ddgan} displays quantitative comparison results between DDGAN and OGDM.
OGDM achieves better FID and recall scores although DDGAN requires fewer steps. 
Note that the performance of DDGAN peaks at $\text{NFEs}=4$ and then drops as the number of time steps increases, which limits the improvement of the model at the expense of increased sampling cost.
Moreover, the low recall scores of DDGAN imply it may share the limitations of GANs such as low diversity and mode collapse.
Also, DDGAN does not support deterministic sampling which makes it hard to solve problems such as inversion.

%

\section{Stochastic sampling}
\label{app:stochastic}

\begin{table*}[h]
    \centering
    \caption{
        FID and recall scores for various NFEs of stochastic sampler where the projection function is Euler method\footref{fn:euler}.
    }
    \label{tab:stochastic}
    \vspace{-2mm}
        \setlength{\tabcolsep}{2.4mm}
    \scalebox{0.85}{
        \begin{tabular}{clcccccc}
    \toprule
    \multicolumn{2}{c}{\diagbox[height=1.2\line]{}{\hspace{15mm} NFEs}} & 
    \multicolumn{2}{c}{50} & \multicolumn{2}{c}{20} & \multicolumn{2}{c}{10} \\ 
    \cmidrule(lr){3-4}\cmidrule(lr){5-6}\cmidrule(lr){7-8}
    Dataset~(Backbone) & Method & FID$\downarrow$ & Rec.$\uparrow$ & FID$\downarrow$ & Rec.$\uparrow$ & FID$\downarrow$ & Rec.$\uparrow$ \\ \hline
    
    \multirow{2}{*}{CIFAR-10~(ADM)}
    & Baseline & 14.28 & 0.491 & 25.42 & 0.388 & 44.37 & 0.278 \\
    & OGDM & \B 9.94 & \B 0.524 & \B 16.84 & \B 0.450 & \B 29.70 & \B 0.359 \\
    \cmidrule(lr){1-8}
    
    \multirow{2}{*}{CelebA~(ADM)}
    & Baseline & 13.51 & 0.312 & 21.00 & 0.181 & 31.09 & 0.079 \\
    & OGDM & \B 9.62 & \B 0.400 & \B 15.74 & \B 0.277 & \B 24.22 & \B 0.158 \\
    \bottomrule    
        \end{tabular}
        }
\end{table*}

\cref{tab:stochastic} presents the quantitative results of stochastic sampling when the projection function is a step of Euler method. 
We observe that the FID and recall are improved for all cases.

\clearpage
\section{Qualitative comparisons}
\label{app:qual}
\begin{figure*}[h]
    \centering
    \setlength{\tabcolsep}{0.5mm}
    \begin{tabular}{cc}
        \includegraphics[width=0.49\textwidth]{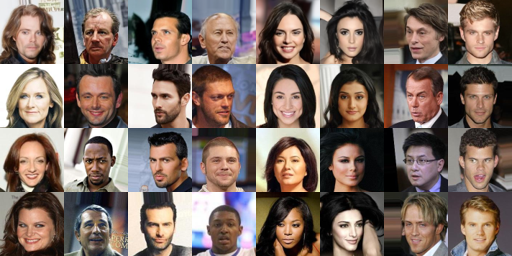} & 
        \includegraphics[width=0.49\textwidth]{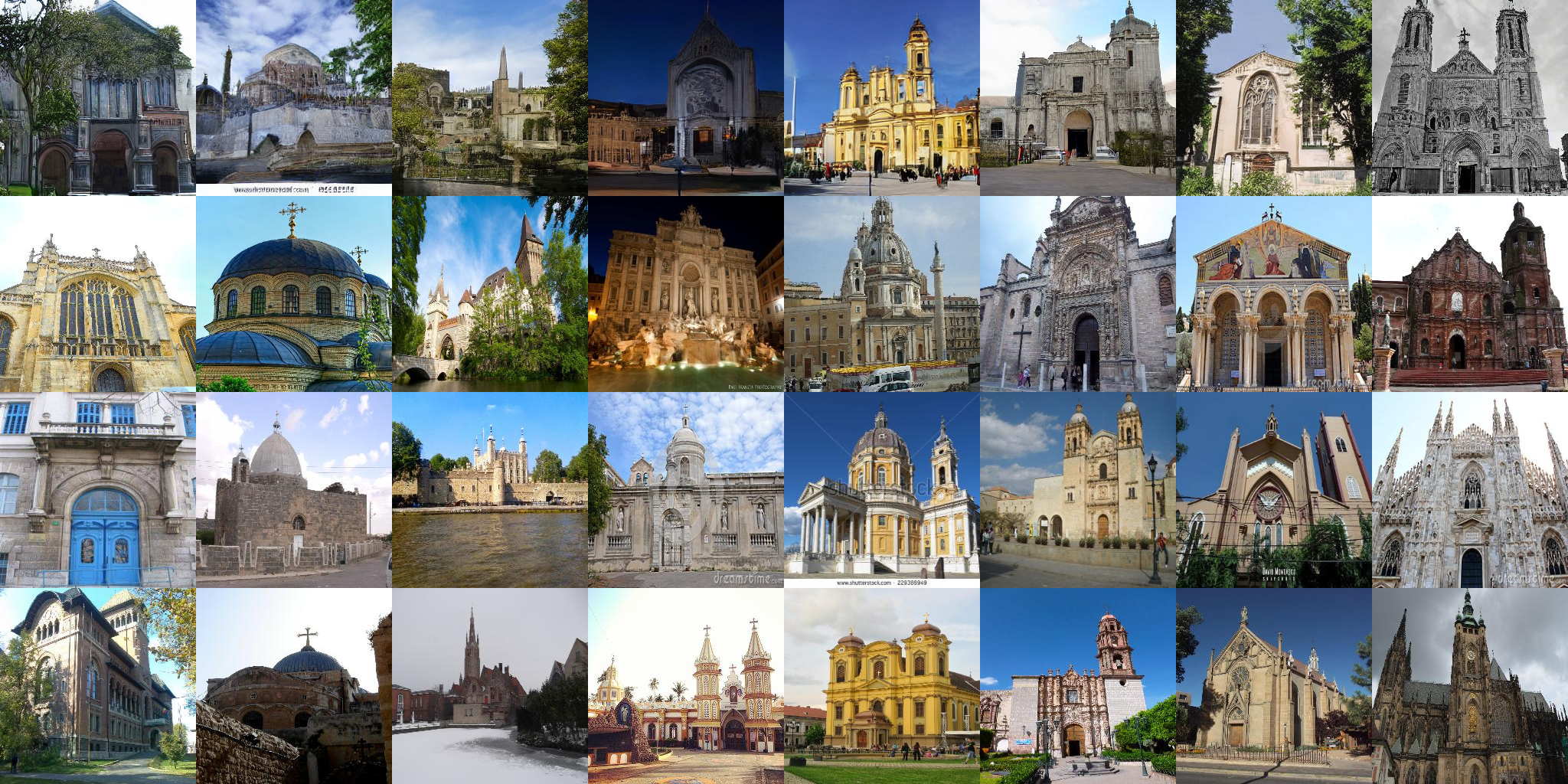} \\
        CelebA  & LSUN Church \\ 
    \end{tabular}
    \caption{
    Nearest neighborhoods of generated samples from CelebA and LSUN Church datasets. The top row showcases our generated samples using Euler method with $\text{NFEs}=50$, while the remaining three rows display the nearest neighborhoods from each training dataset. 
    The distances are measured in the Inception-v3~\cite{szegedy2016rethinking} feature space. 
	}
    \label{fig:nn}  
\end{figure*}

We compare the generated images between the baseline and our method using few number of NFEs in \cref{fig:qual_cifar10_nfe10,fig:qual_cifar10_nfe15,fig:qual_cifar10_nfe20,fig:qual_cifar10_nfe25,fig:qual_cifar10_edm_nfe10,fig:qual_cifar10_edm_nfe15,fig:qual_cifar10_edm_nfe20,fig:qual_cifar10_edm_nfe25,fig:qual_celeba_nfe10,fig:qual_celeba_nfe15,fig:qual_celeba_nfe20,fig:qual_celeba_nfe25,fig:qual_lsun_church_nfe10,fig:qual_lsun_church_nfe15,fig:qual_lsun_church_nfe20,fig:qual_lsun_church_nfe25}.
While we use Euler method and PNDM for sampling in common, for CIFAR-10, we further compare the results on EDM backbone sampled by Heun's method.
The images generated by our method have more vivid color and clearer and less prone to produce unrealistic samples compared to the baselines.
Also, our method complements advanced samplers, other than Euler method, effectively.
In addition, \cref{fig:nn} in \cref{app:nn} illustrates the nearest neighbor examples of our generated examples in training data of CelebA and LSUN Church.

\subsection{CIFAR-10 samples with ADM baseline}
\label{app:qual_cifar10_adm}
\begin{figure*}[h!]
    \centering
    \setlength{\tabcolsep}{0.5mm}
    \begin{tabular}{cc}
        \includegraphics[width=0.49\textwidth]{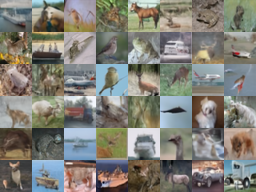} & 
        \includegraphics[width=0.49\textwidth]{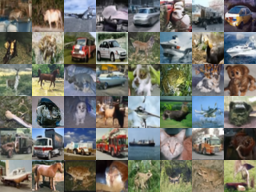} \\
        Baseline + Euler method (NFEs=$10$) & OGDM + Euler method (NFEs=$10$)\\ 
        (FID: 15.20, recall: 0.527) & (FID: 11.18, recall: 0.549) \\
        \includegraphics[width=0.49\textwidth]{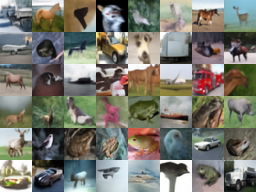} & 
        \includegraphics[width=0.49\textwidth]{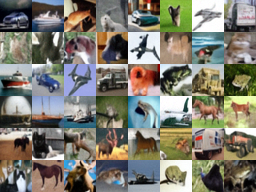} \\
        Baseline + S-PNDM (NFEs$=10$)& OGDM + S-PNDM (NFEs$=10$)\\
        (FID: 10.32, recall: 0.548) & (FID: 7.54, recall: 0.582)
    \end{tabular}
    \caption{
    Qualitative results on CIFAR-10 dataset with the ADM backbone using Euler method (top) and S-PNDM (bottom) with NFEs=$10$.
	}
    \label{fig:qual_cifar10_nfe10}  
\end{figure*}
\begin{figure*}[th!]
    \centering
    \setlength{\tabcolsep}{0.5mm}
    \begin{tabular}{cc}
        \includegraphics[width=0.49\textwidth]{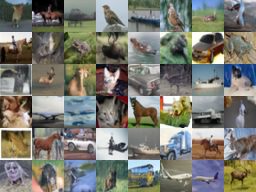} & 
        \includegraphics[width=0.49\textwidth]{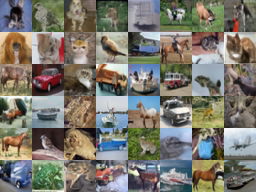} \\
        Baseline + Euler method (NFEs=$15$) & OGDM + Euler method (NFEs=$15$)\\ 
        (FID: 9.93, recall: 0.567) & (FID: 7.96, recall: 0.578) \\
        \includegraphics[width=0.49\textwidth]{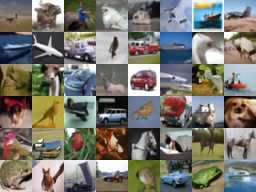} & 
        \includegraphics[width=0.49\textwidth]{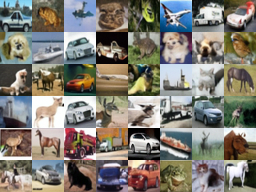} \\
        Baseline + S-PNDM (NFEs$=15$)& OGDM + S-PNDM (NFEs$=15$)\\
        (FID: 7.09, recall: 0.577) & (FID: 5.58, recall: 0.601)
    \end{tabular}
    \caption{
    Qualitative results on CIFAR-10 dataset with the ADM backbone using Euler method (top) and S-PNDM (bottom) with NFEs=$15$.
	}
    \label{fig:qual_cifar10_nfe15}  
\end{figure*}
\begin{figure*}[th!]
    \centering
    \setlength{\tabcolsep}{0.5mm}
    \begin{tabular}{cc}
        \includegraphics[width=0.49\textwidth]{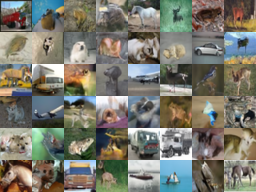} & 
        \includegraphics[width=0.49\textwidth]{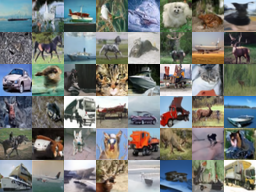} \\
        Baseline + Euler method (NFEs=$20$) & OGDM + Euler method (NFEs=$20$)\\ 
        (FID: 8.05, recall: 0.582) & (FID: 6.81, recall: 0.587) \\
        \includegraphics[width=0.49\textwidth]{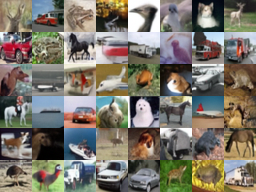} & 
        \includegraphics[width=0.49\textwidth]{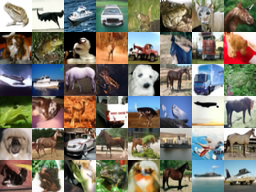} \\
        Baseline + S-PNDM (NFEs$=20$)& OGDM + S-PNDM (NFEs$=20$)\\
        (FID: 5.95, recall: 0.596) & (FID: 5.09, recall: 0.605)
    \end{tabular}
    \caption{
    Qualitative results on CIFAR-10 dataset with the ADM backbone using Euler method (top) and S-PNDM (bottom) with NFEs=$20$.
	}
    \label{fig:qual_cifar10_nfe20}  
\end{figure*}
\begin{figure*}[th!]
    \centering
    \setlength{\tabcolsep}{0.5mm}
    \begin{tabular}{cc}
        \includegraphics[width=0.49\textwidth]{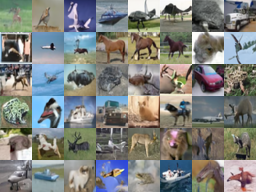} & 
        \includegraphics[width=0.49\textwidth]{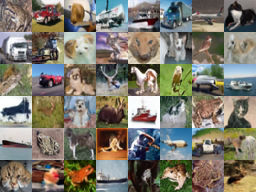} \\
        Baseline + Euler method (NFEs=$25$) & OGDM + Euler method (NFEs=$25$)\\ 
        (FID: 7.08, recall: 0.583) & (FID: 6.26, recall: 0.587) \\
        \includegraphics[width=0.49\textwidth]{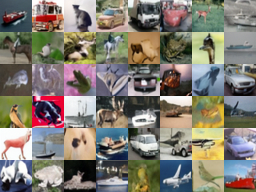} & 
        \includegraphics[width=0.49\textwidth]{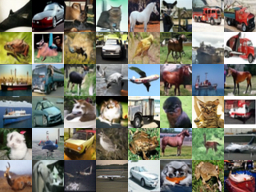} \\
        Baseline + S-PNDM (NFEs$=25$) & OGDM + S-PNDM (NFEs$=25$)\\
        (FID: 5.31, recall: 0.601) & (FID: 5.03, recall: 0.611)
    \end{tabular}
    \caption{
    Qualitative results on CIFAR-10 dataset with the ADM backbone using Euler method (top) and S-PNDM (bottom) with NFEs=$25$.
	}
    \label{fig:qual_cifar10_nfe25}  
\end{figure*}
%


\clearpage
\subsection{CIFAR-10 samples with EDM baseline}
\label{app:qual_cifar10_edm}
\begin{figure*}[th!]
    \centering
    \setlength{\tabcolsep}{0.5mm}
    \begin{tabular}{cc}
        \includegraphics[width=0.49\textwidth]{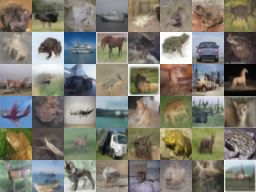} & 
        \includegraphics[width=0.49\textwidth]{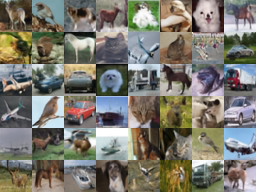} \\
        Baseline + Euler method (NFEs=$10$)& OGDM + Euler method (NFEs=$10$)\\ 
        (FID: 19.32, recall: 0.452) & (FID: 9.28, recall: 0.546) \\
        \includegraphics[width=0.49\textwidth]{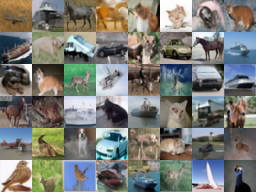} & 
        \includegraphics[width=0.49\textwidth]{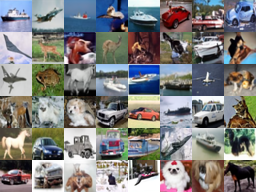} \\
        Baseline + S-PNDM (NFEs=$10$)& OGDM + S-PNDM (NFEs=$10$)\\ 
        (FID: 9.51, recall: 0.531) & (FID: 4.97, recall: 0.586) \\
    \end{tabular}
    \caption{
    Qualitative results on CIFAR-10 dataset with the EDM backbone using Euler method (top), and S-PNDM (bottom) with NFEs=$10$.
	}
    \label{fig:qual_cifar10_edm_nfe10}  
\end{figure*}
\begin{figure*}[th!]
    \centering
    \setlength{\tabcolsep}{0.5mm}
    \begin{tabular}{cc}
        \includegraphics[width=0.49\textwidth]{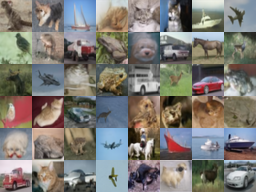} & 
        \includegraphics[width=0.49\textwidth]{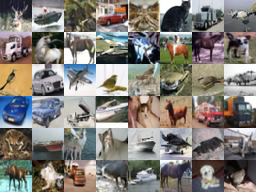} \\
        Baseline + Euler method (NFEs=$15$)& OGDM + Euler method (NFEs=$15$)\\ 
        (FID: 10.02, recall: 0.524) & (FID: 4.64, recall: 0.578) \\
        \includegraphics[width=0.49\textwidth]{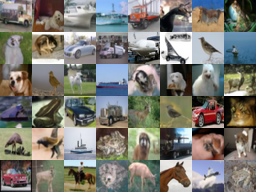} & 
        \includegraphics[width=0.49\textwidth]{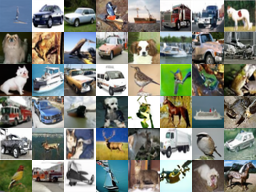} \\
        Baseline + S-PNDM (NFEs=$15$)& OGDM + S-PNDM (NFEs=$15$)\\ 
        (FID: 4.48, recall: 0.586) & (FID: 3.60, recall: 0.600) \\
        \includegraphics[width=0.49\textwidth]{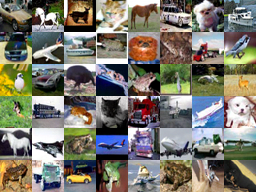} & 
        \includegraphics[width=0.49\textwidth]{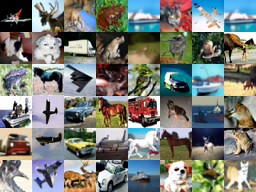} \\
        Baseline + Heun's method (NFEs=$15$)& OGDM + Heun's method (NFEs=$15$)\\ 
        (FID: 4.48, recall: 0.604) & (FID: 4.21, recall: 0.619)
    \end{tabular}
    \caption{
    Qualitative results on CIFAR-10 dataset with the EDM backbone using Euler method (top), S-PNDM (middle) and Heun's method (bottom) with NFEs=$15$.
	}
    \label{fig:qual_cifar10_edm_nfe15}  
\end{figure*}
\begin{figure*}[th!]
    \centering
    \setlength{\tabcolsep}{0.5mm}
    \begin{tabular}{cc}
        \includegraphics[width=0.49\textwidth]{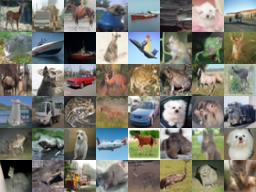} & 
        \includegraphics[width=0.49\textwidth]{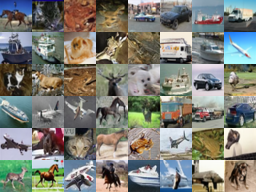} \\
        Baseline + Euler method (NFEs=$20$)& OGDM + Euler method (NFEs=$20$)\\ 
        (FID: 6.82, recall: 0.558) & (FID: 3.53, recall: 0.600) \\
        \includegraphics[width=0.49\textwidth]{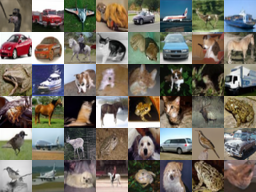} & 
        \includegraphics[width=0.49\textwidth]{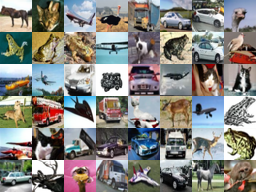} \\
        Baseline + S-PNDM (NFEs=$20$)& OGDM + S-PNDM (NFEs=$20$)\\ 
        (FID: 3.21, recall: 0.597) & (FID: 3.62, recall: 0.605) \\
    \end{tabular}
    \caption{
    Qualitative results on CIFAR-10 dataset with the EDM backbone using Euler method (top), and S-PNDM (bottom) with NFEs=$20$.
	}
    \label{fig:qual_cifar10_edm_nfe20}  
\end{figure*}
\begin{figure*}[th!]
    \centering
    \setlength{\tabcolsep}{0.5mm}
    \begin{tabular}{cc}
        \includegraphics[width=0.49\textwidth]{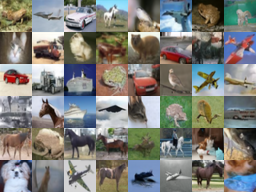} & 
        \includegraphics[width=0.49\textwidth]{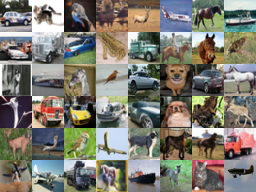} \\
        Baseline + Euler method (NFEs=$25$)& OGDM + Euler method (NFEs=$25$)\\ 
        (FID: 5.32, recall: 0.572) & (FID: 3.21, recall: 0.603) \\
        \includegraphics[width=0.49\textwidth]{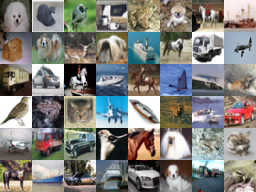} & 
        \includegraphics[width=0.49\textwidth]{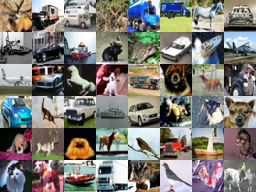} \\
        Baseline + S-PNDM (NFEs=$25$)& OGDM + S-PNDM (NFEs=$25$)\\ 
        (FID: 2.74, recall: 0.604) & (FID: 3.75, recall: 0.604) \\
        \includegraphics[width=0.49\textwidth]{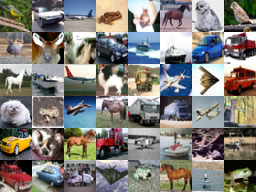} & 
        \includegraphics[width=0.49\textwidth]{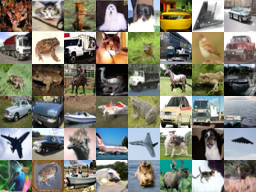} \\
        Baseline + Heun's method (NFEs=$25$)& OGDM + Heun's method (NFEs=$25$)\\ 
        (FID: 2.19, recall: 0.616) & (FID: 2.17, recall: 0.622)
    \end{tabular}
    \caption{
    Qualitative results on CIFAR-10 dataset with the EDM backbone using Euler method (top), S-PNDM (middle) and Heun's method (bottom) with NFEs=$25$.
	}
    \label{fig:qual_cifar10_edm_nfe25}  
\end{figure*}
%

\clearpage
\subsection{CelebA samples with ADM baseline}
\label{app:qual_celeba_adm}
\begin{figure*}[th!]
    \centering
    \setlength{\tabcolsep}{0.5mm}
    \begin{tabular}{cc}
        \includegraphics[width=0.49\textwidth]{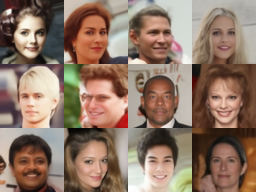} & 
        \includegraphics[width=0.49\textwidth]{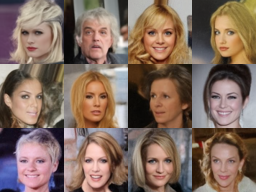} \\
        Baseline + Euler method (NFEs=$10$) & OGDM + Euler method (NFEs=$10$)\\ 
        (FID: 11.92, recall: 0.315) & (FID: 7.04, recall: 0.504)\\
        \includegraphics[width=0.49\textwidth]{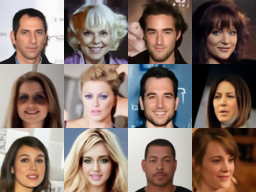} & 
        \includegraphics[width=0.49\textwidth]{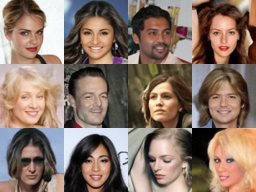} \\
        Baseline + S-PNDM (NFEs=$10$) & OGDM + S-PNDM (NFEs=$10$)\\ 
        (FID: 7.33, recall: 0.445) & (FID: 4.35, recall: 0.545)
    \end{tabular}
    \caption{
        Qualitative results on CelebA dataset with the ADM backbone using Euler method (top) and S-PNDM (bottom) with NFEs=$10$.
    }
    \label{fig:qual_celeba_nfe10}  
\end{figure*}
\begin{figure*}[th!]
    \centering
    \setlength{\tabcolsep}{0.5mm}
    \begin{tabular}{cc}
        \includegraphics[width=0.49\textwidth]{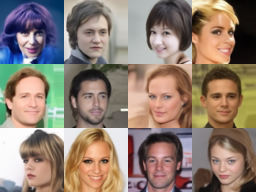} & 
        \includegraphics[width=0.49\textwidth]{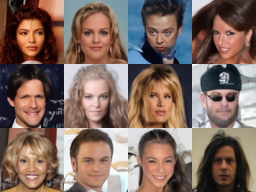} \\
        Baseline + Euler method (NFEs=$15$) & OGDM + Euler method (NFEs=$15$)\\ 
        (FID: 9.34, recall: 0.392) & (FID: 4.80, recall: 0.552)\\
        \includegraphics[width=0.49\textwidth]{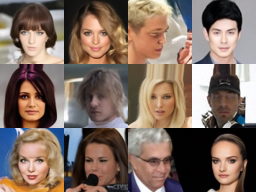} & 
        \includegraphics[width=0.49\textwidth]{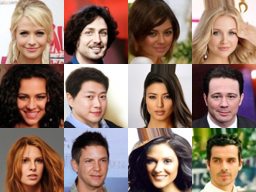} \\
        Baseline + S-PNDM (NFEs=$15$) & OGDM + S-PNDM (NFEs=$15$)\\ 
        (FID: 5.22, recall: 0.511) & (FID: 2.96, recall: 0.585)
    \end{tabular}
    \caption{
        Qualitative results on CelebA dataset with the ADM backbone using Euler method (top) and S-PNDM (bottom) with NFEs=$15$.
    }
    \label{fig:qual_celeba_nfe15}  
\end{figure*}
\begin{figure*}[th!]
    \centering
    \setlength{\tabcolsep}{0.5mm}
    \begin{tabular}{cc}
        \includegraphics[width=0.49\textwidth]{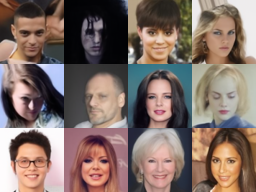} & 
        \includegraphics[width=0.49\textwidth]{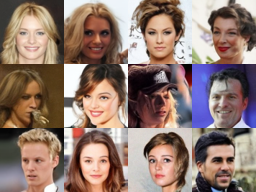} \\
        Baseline + Euler method (NFEs=$20$) & OGDM + Euler method (NFEs=$20$)\\ 
        (FID: 7.88, recall: 0.429) & (FID: 3.94, recall: 0.534)\\
        \includegraphics[width=0.49\textwidth]{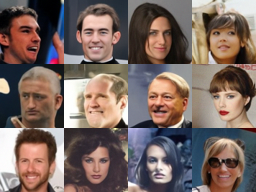} & 
        \includegraphics[width=0.49\textwidth]{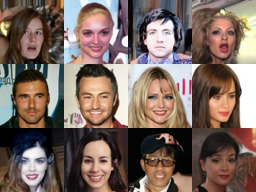} \\
        Baseline + S-PNDM (NFEs=$20$) & OGDM + S-PNDM (NFEs=$20$)\\ 
        (FID: 4.15, recall: 0.540) & (FID: 2.70, recall: 0.604)
    \end{tabular}
    \caption{
        Qualitative results on CelebA dataset with the ADM backbone using Euler method (top) and S-PNDM (bottom) with NFEs=$20$.
    }
    \label{fig:qual_celeba_nfe20}  
\end{figure*}
\begin{figure*}[th!]
    \centering
    \setlength{\tabcolsep}{0.5mm}
    \begin{tabular}{cc}
        \includegraphics[width=0.49\textwidth]{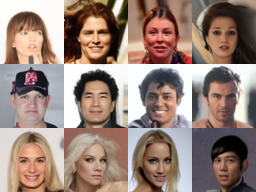} & 
        \includegraphics[width=0.49\textwidth]{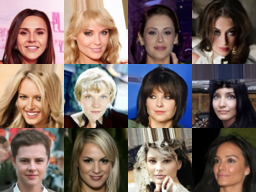} \\
        Baseline + Euler method (NFEs=$25$) & OGDM + Euler method (NFEs=$25$)\\ 
        (FID: 7.20, recall: 0.441) & (FID: 3.80, recall: 0.541)\\
        \includegraphics[width=0.49\textwidth]{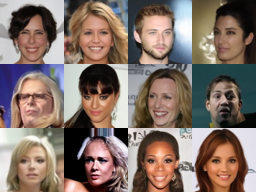} & 
        \includegraphics[width=0.49\textwidth]{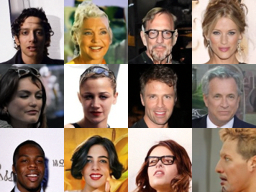} \\
        Baseline + S-PNDM (NFEs=$25$) & OGDM + S-PNDM (NFEs=$25$)\\ 
        (FID: 3.67, recall: 0.553) & (FID: 2.62, recall: 0.607)
    \end{tabular}
    \caption{
        Qualitative results on CelebA dataset with the ADM backbone using Euler method (top) and S-PNDM (bottom) with NFEs=$25$.
    }
    \label{fig:qual_celeba_nfe25}  
\end{figure*}

\clearpage
\subsection{LSUN Church samples with LDM baseline}
\label{app:qual_lsun_ldm}
\begin{figure*}[th!]
    \centering
    \setlength{\tabcolsep}{0.3mm}
    \begin{tabular}{cc}
        \includegraphics[width=0.49\textwidth]{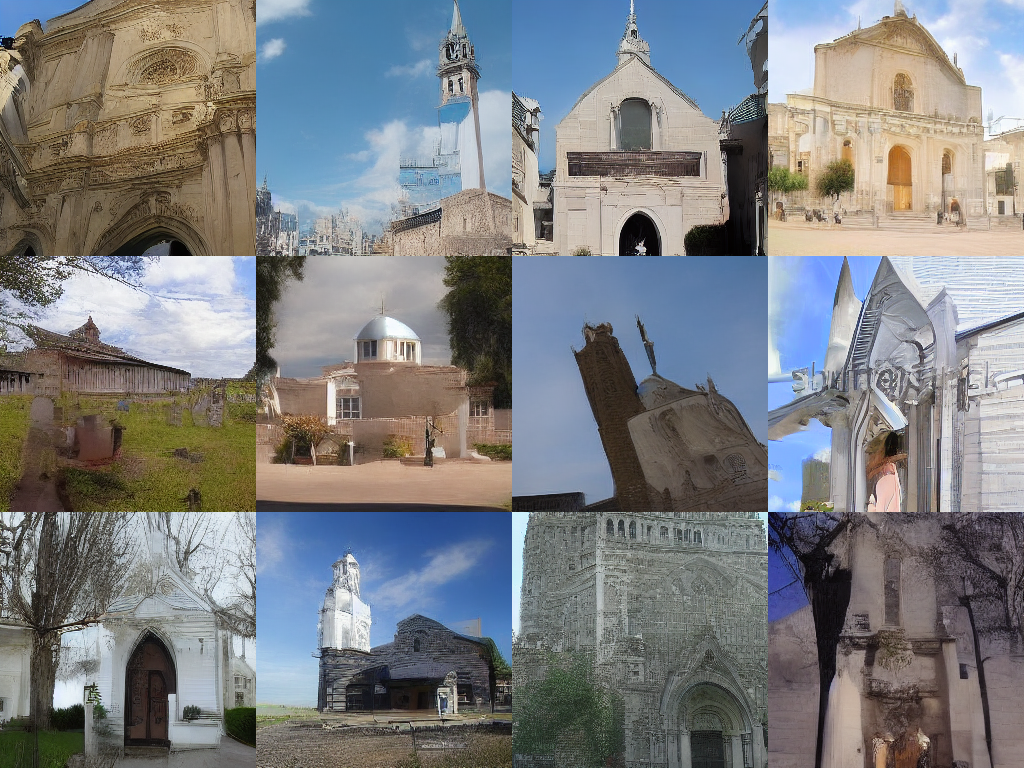} & 
        \includegraphics[width=0.49\textwidth]{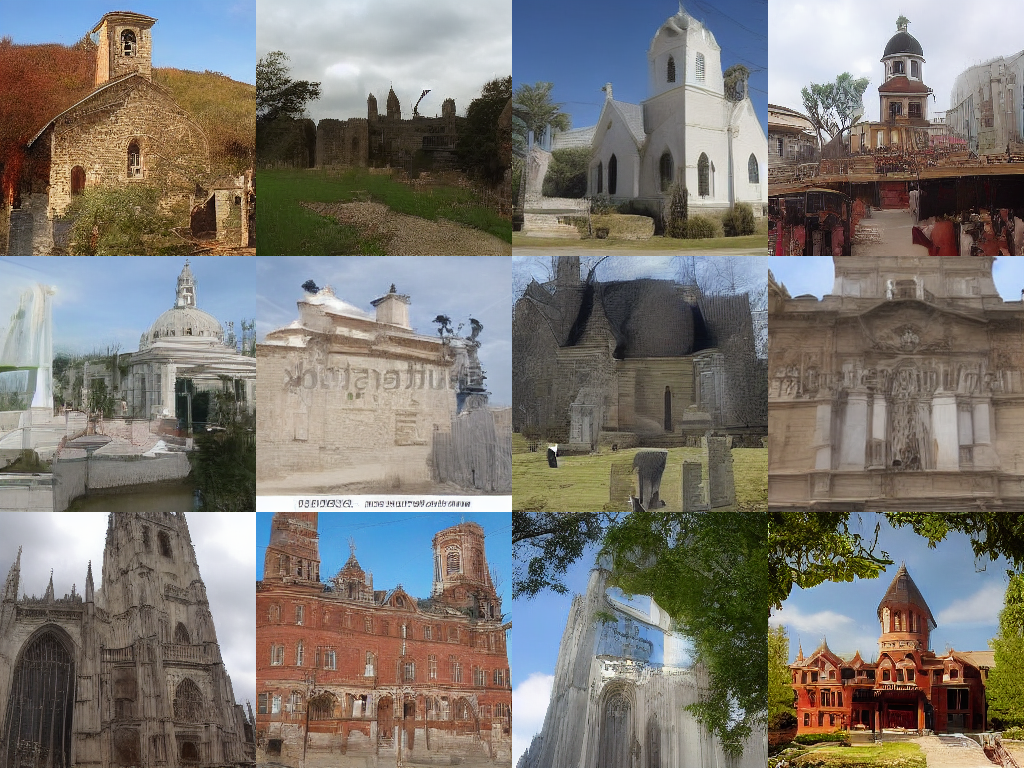} \\
        Baseline + Euler method (NFEs=$10$) & OGDM + Euler method (NFEs=$10$)\\ 
        (FID: 15.02, recall: 0.326) & (FID: 14.84, recall: 0.331) \\
        \includegraphics[width=0.49\textwidth]{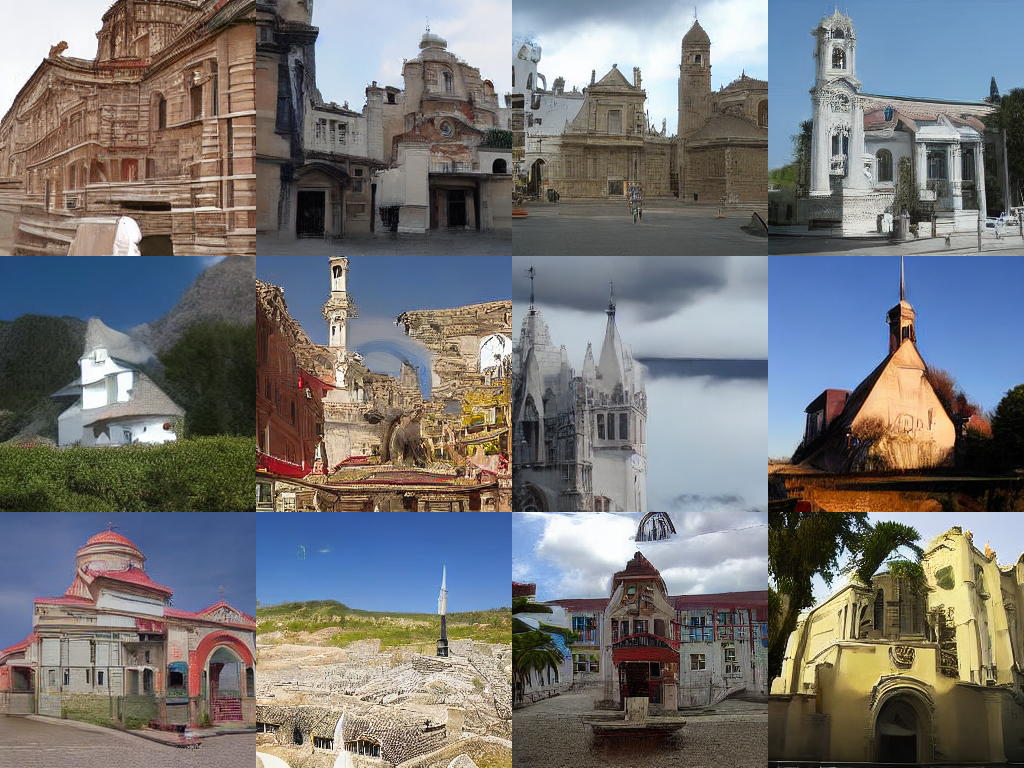} & 
        \includegraphics[width=0.49\textwidth]{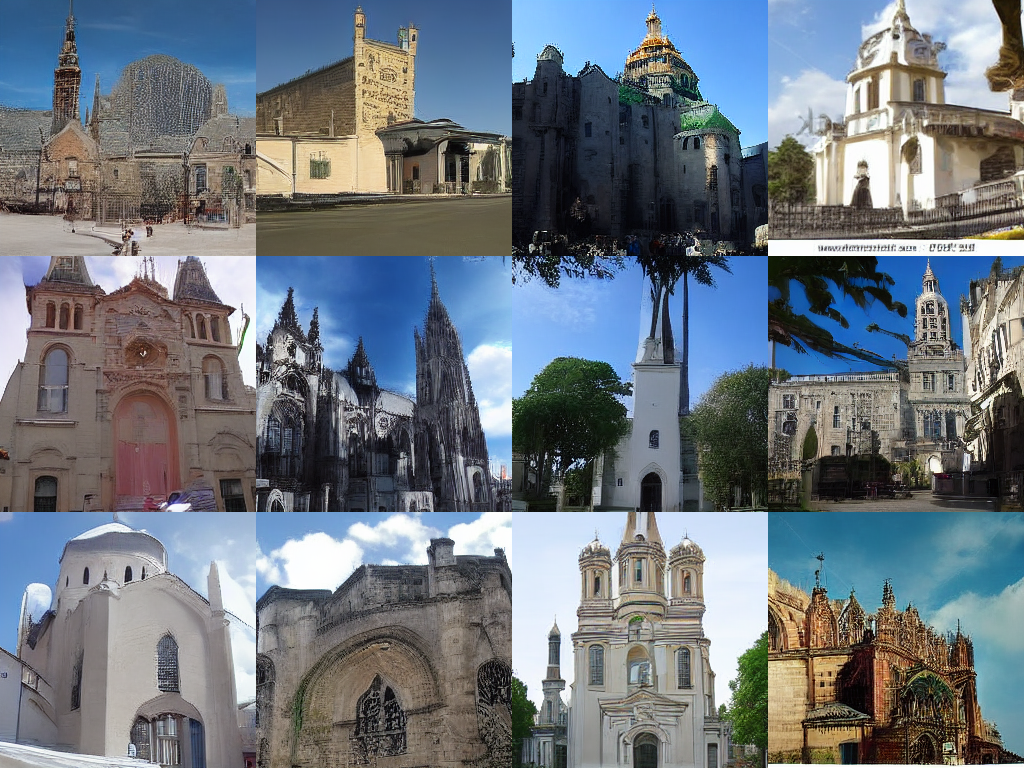} \\
        Baseline + S-PNDM (NFEs$=10$) & OGDM + S-PNDM (NFEs$=10$) \\
        (FID: 9.14, recall: 0.464) & (FID: 8.68, recall: 0.478)
    \end{tabular}
    \caption{
        Qualitative results on LSUN Church dataset with the LDM backbone using Euler method (top) and S-PNDM (bottom) with NFEs=$10$.
    }
    \label{fig:qual_lsun_church_nfe10}  
\end{figure*}
\begin{figure*}[th!]
    \centering
    \setlength{\tabcolsep}{0.3mm}
    \begin{tabular}{cc}
        \includegraphics[width=0.49\textwidth]{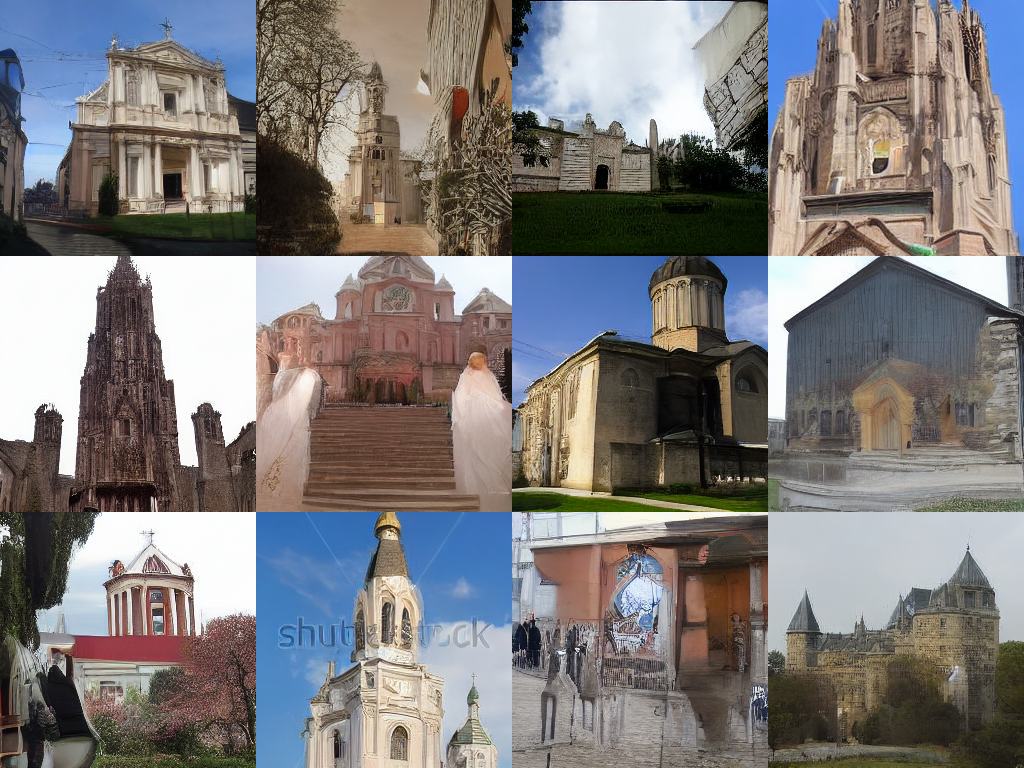} & 
        \includegraphics[width=0.49\textwidth]{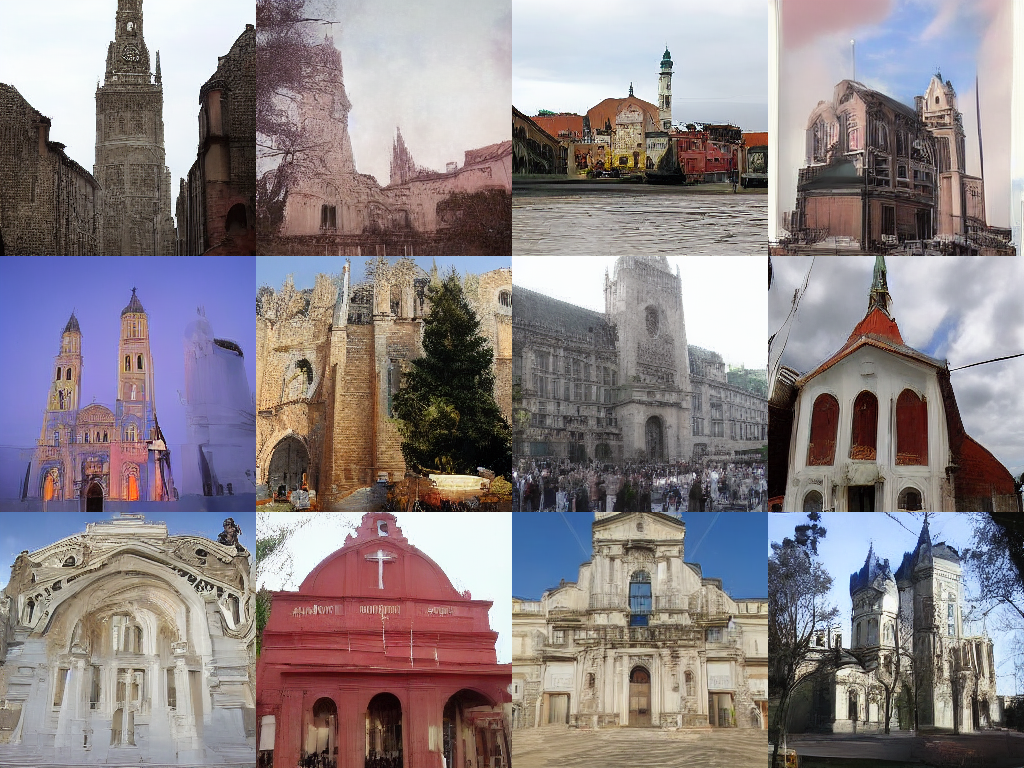} \\
        Baseline + Euler method (NFEs=$15$) & OGDM + Euler method (NFEs=$15$)\\ 
        (FID: 8.83, recall: 0.399) & (FID: 8.76, recall: 0.402) \\
        \includegraphics[width=0.49\textwidth]{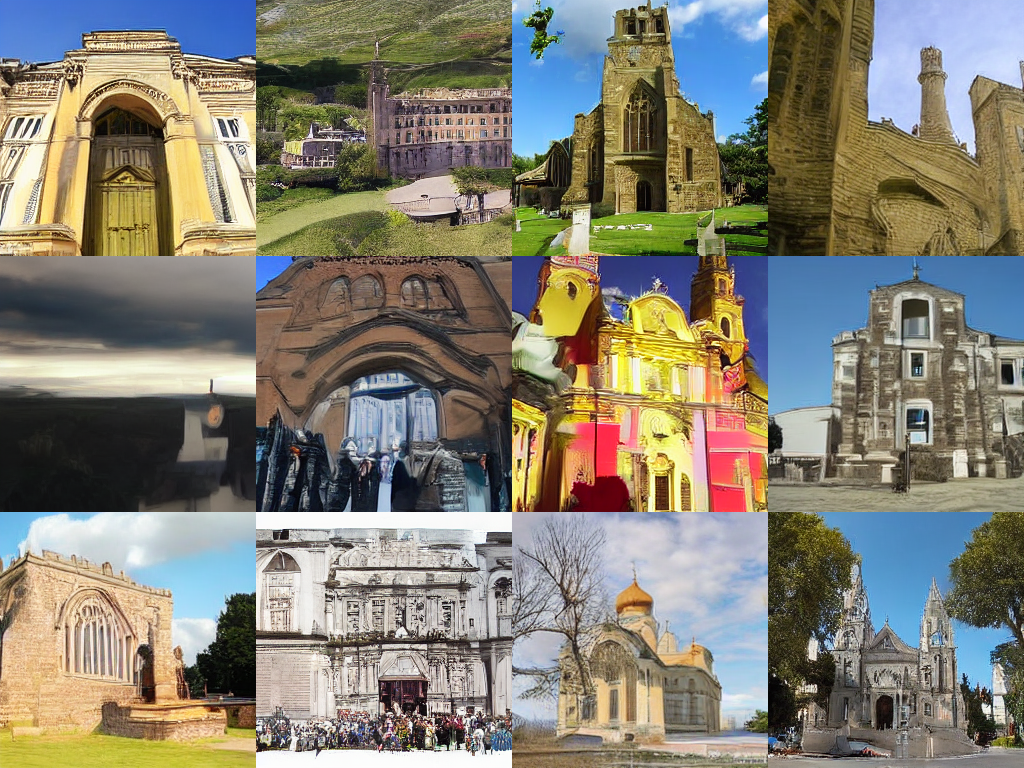} & 
        \includegraphics[width=0.49\textwidth]{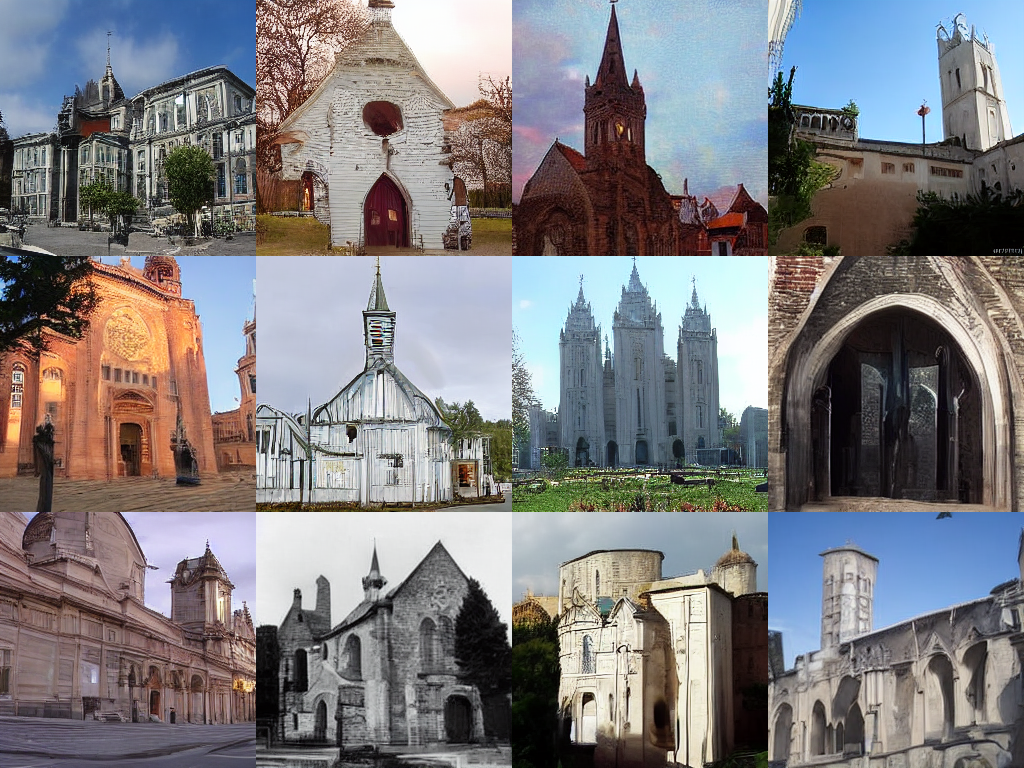} \\
        Baseline + S-PNDM (NFEs$=15$) & OGDM + S-PNDM (NFEs$=15$) \\
        (FID: 8.07, recall: 0.475) & (FID: 7.48, recall: 0.481) \\
        \includegraphics[width=0.49\textwidth]{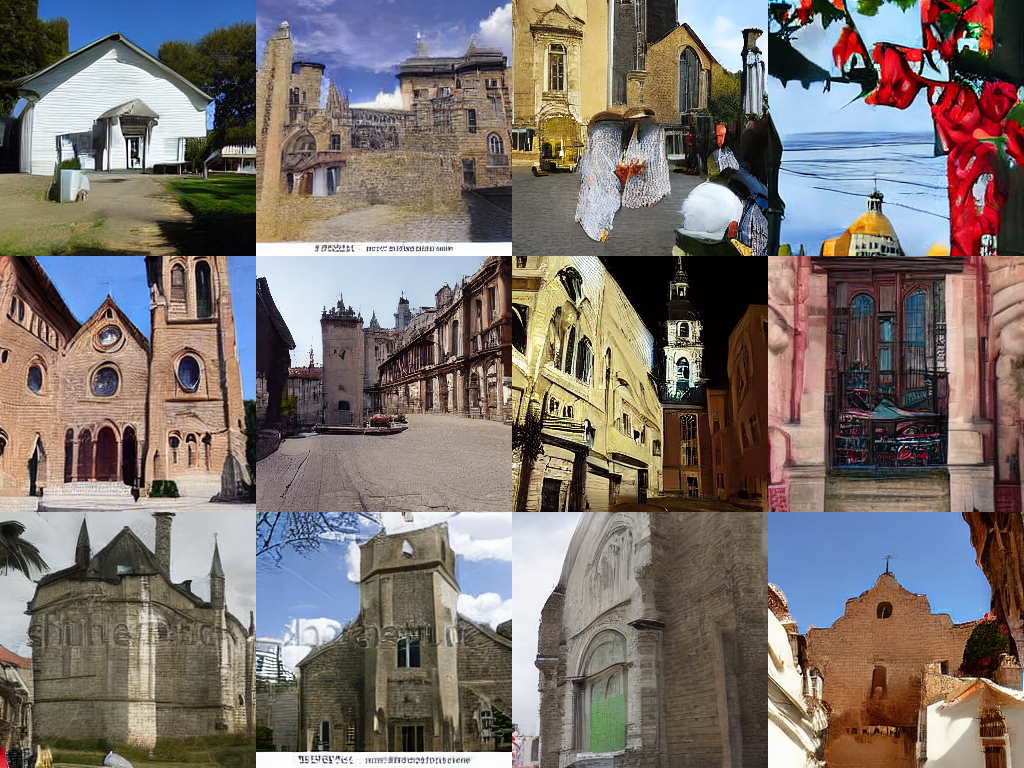} & 
        \includegraphics[width=0.49\textwidth]{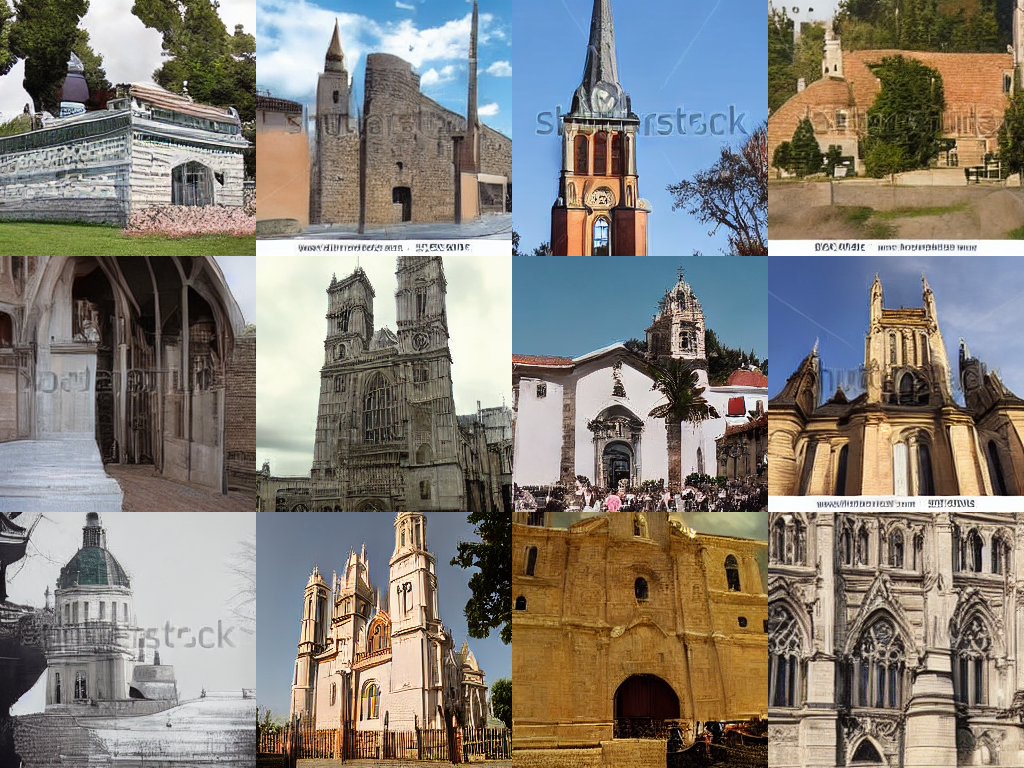} \\
        Baseline + F-PNDM (NFEs$=15$) & OGDM + F-PNDM (NFEs$=15$) \\
        (FID: 12.75, recall: 0.493) & (FID: 11.78, recall: 0.505)
    \end{tabular}
    \caption{
        Qualitative results on LSUN Church dataset with the LDM backbone using Euler method (top), S-PNDM (middle) and F-PNDM (bottom) with NFEs=$15$.
    }
    \label{fig:qual_lsun_church_nfe15}  
\end{figure*}
\begin{figure*}[th!]
    \centering
    \setlength{\tabcolsep}{0.3mm}
    \begin{tabular}{cc}
        \includegraphics[width=0.49\textwidth]{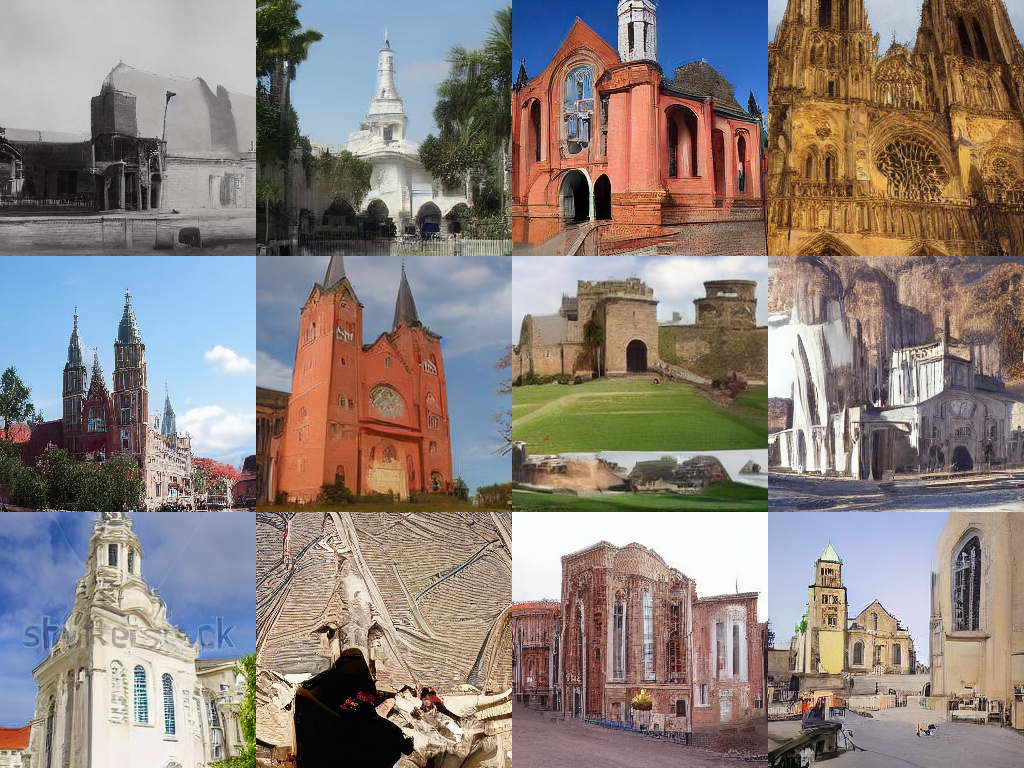} & 
        \includegraphics[width=0.49\textwidth]{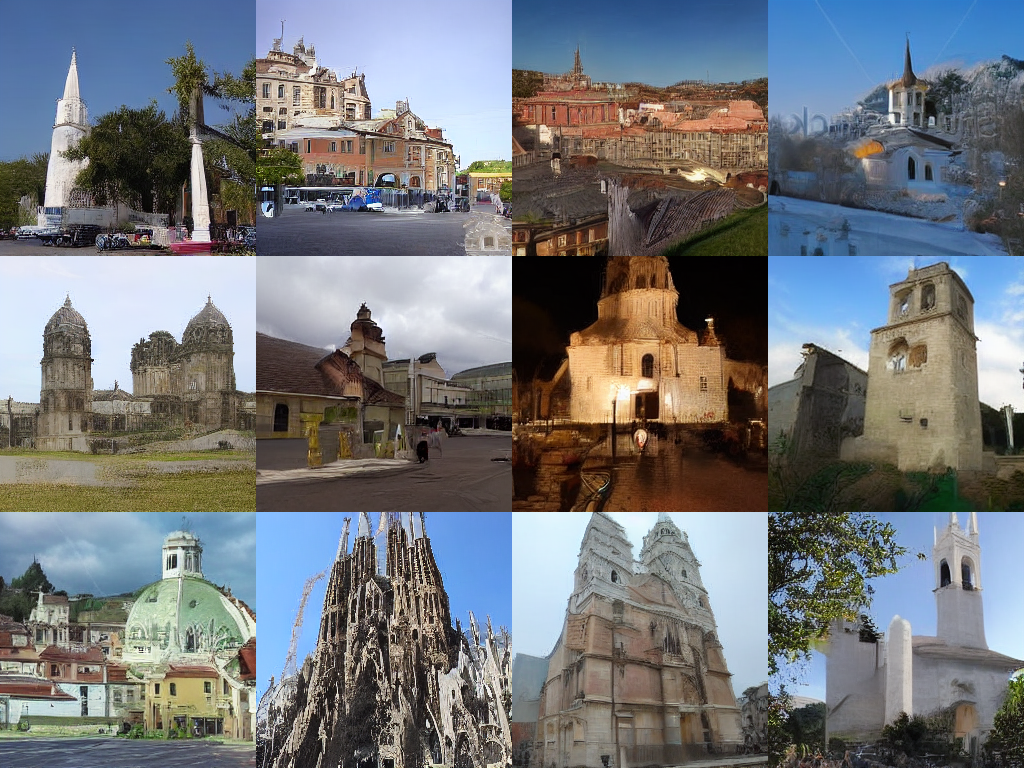} \\
        Baseline + Euler method (NFEs=$20$) & OGDM + Euler method (NFEs=$20$)\\ 
        (FID: 8.40, recall: 0.434) & (FID: 7.92, recall: 0.444) \\
        \includegraphics[width=0.49\textwidth]{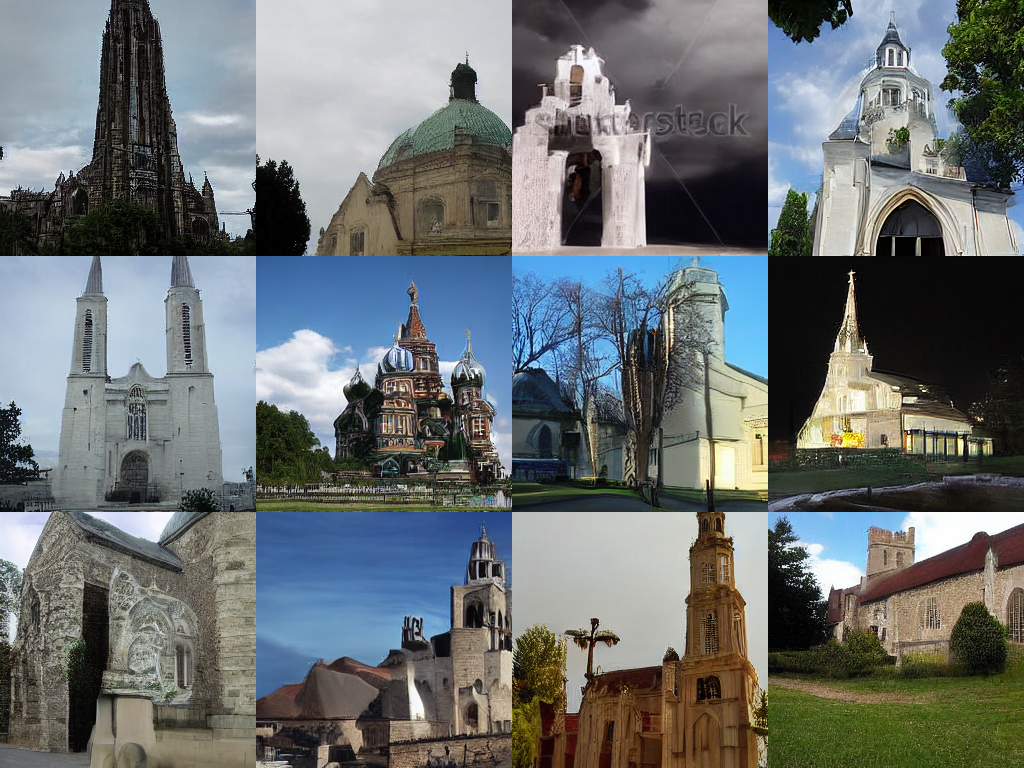} & 
        \includegraphics[width=0.49\textwidth]{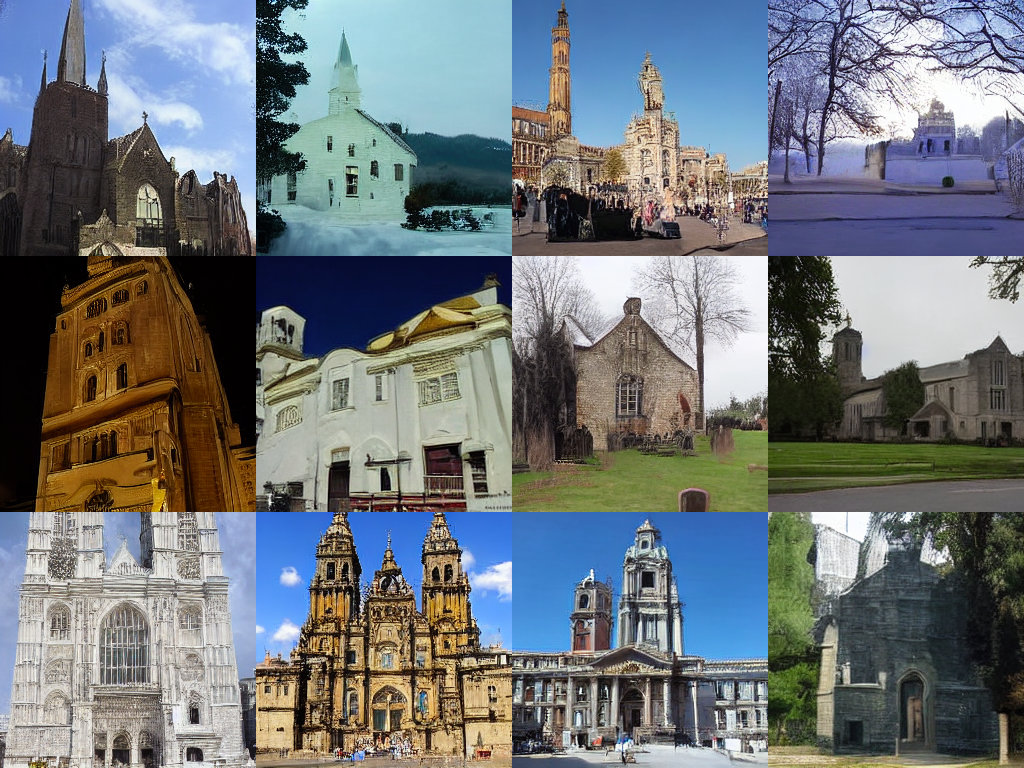} \\
        Baseline + S-PNDM (NFEs$=20$) & OGDM + S-PNDM (NFEs$=20$) \\
        (FID: 8.21, recall: 0.471) & (FID: 7.48, recall: 0.489) \\
        \includegraphics[width=0.49\textwidth]{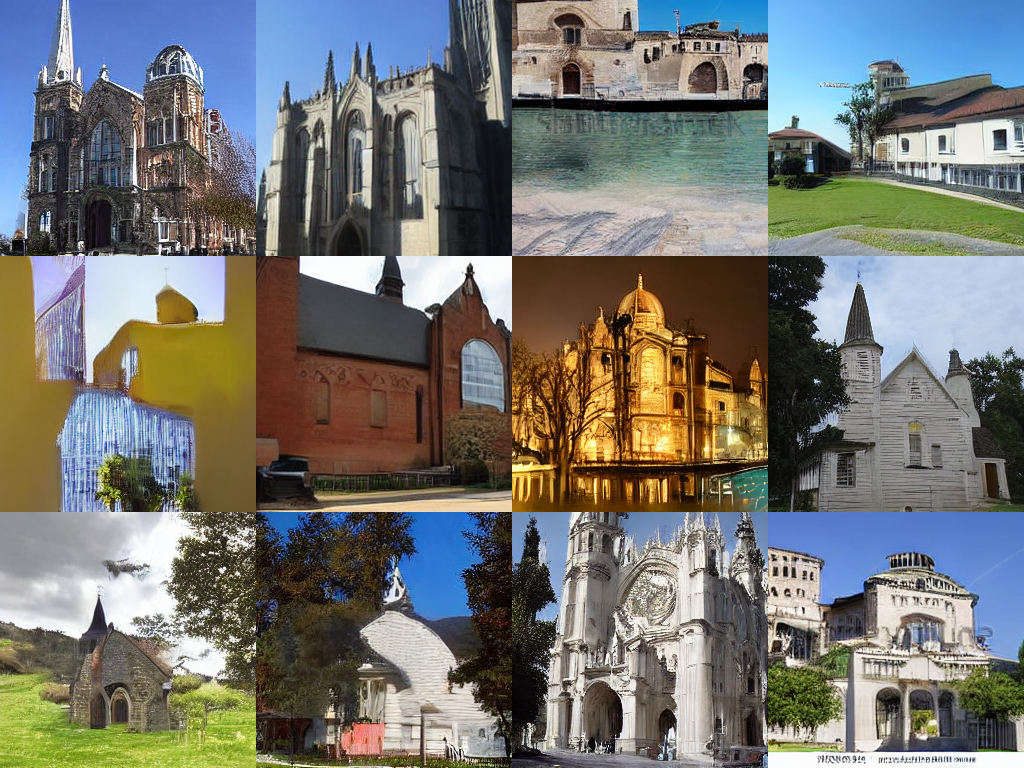} & 
        \includegraphics[width=0.49\textwidth]{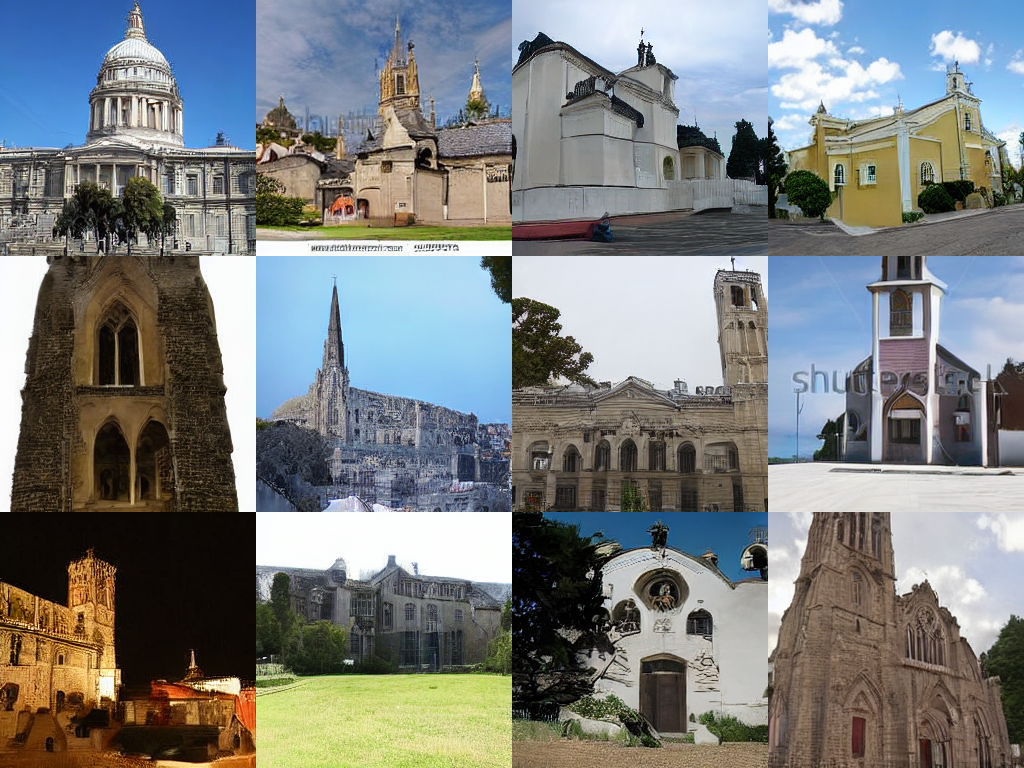} \\
        Baseline + F-PNDM (NFEs$=20$) & OGDM + F-PNDM (NFEs$=20$) \\
        (FID: 9.10, recall: 0.483) & (FID: 8.39, recall: 0.495)
    \end{tabular}
    \caption{
        Qualitative results on LSUN Church dataset with the LDM backbone using Euler method (top), S-PNDM (middle) and F-PNDM (bottom) with NFEs=$20$.
    }
    \label{fig:qual_lsun_church_nfe20}  
\end{figure*}
\begin{figure*}[th!]
    \centering
    \setlength{\tabcolsep}{0.3mm}
    \begin{tabular}{cc}
        \includegraphics[width=0.49\textwidth]{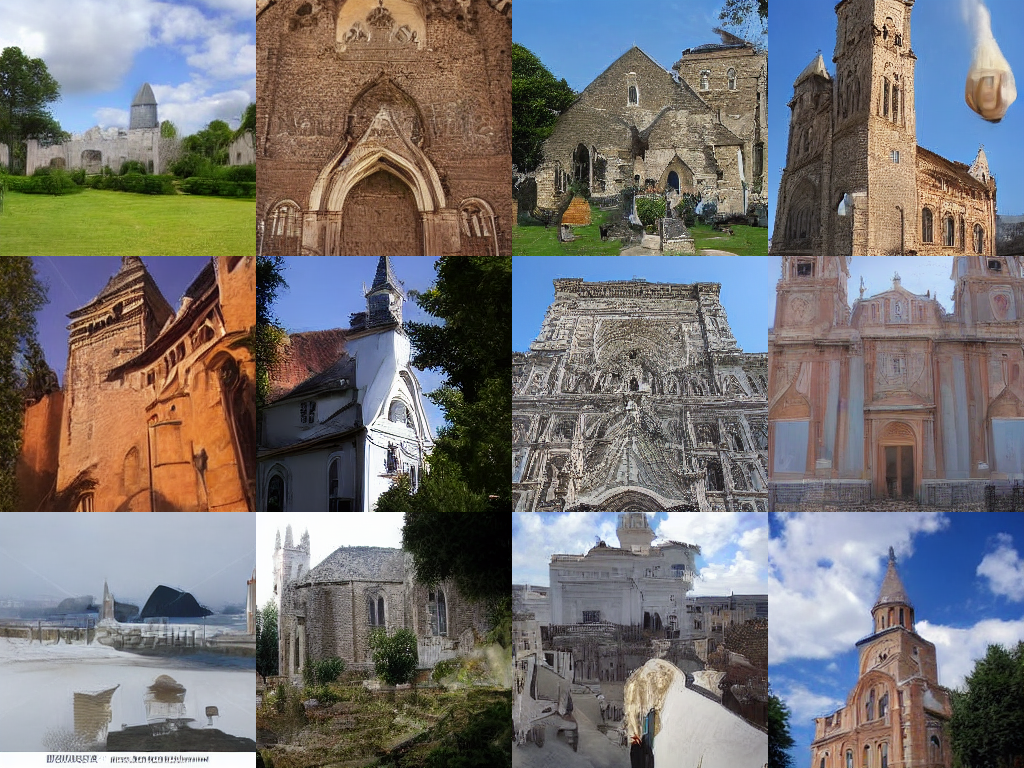} & 
        \includegraphics[width=0.49\textwidth]{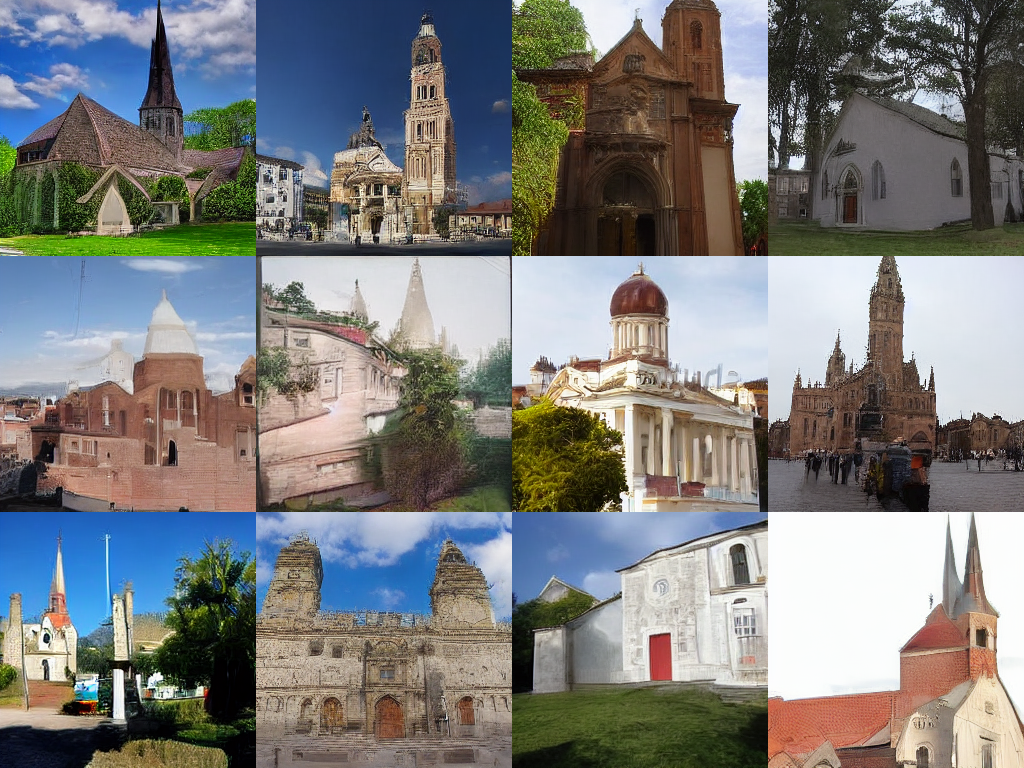} \\
        Baseline + Euler method (NFEs=$25$) & OGDM + Euler method (NFEs=$25$)\\ 
        (FID: 7.87, recall: 0.443) & (FID: 7.46, recall: 0.449) \\
        \includegraphics[width=0.49\textwidth]{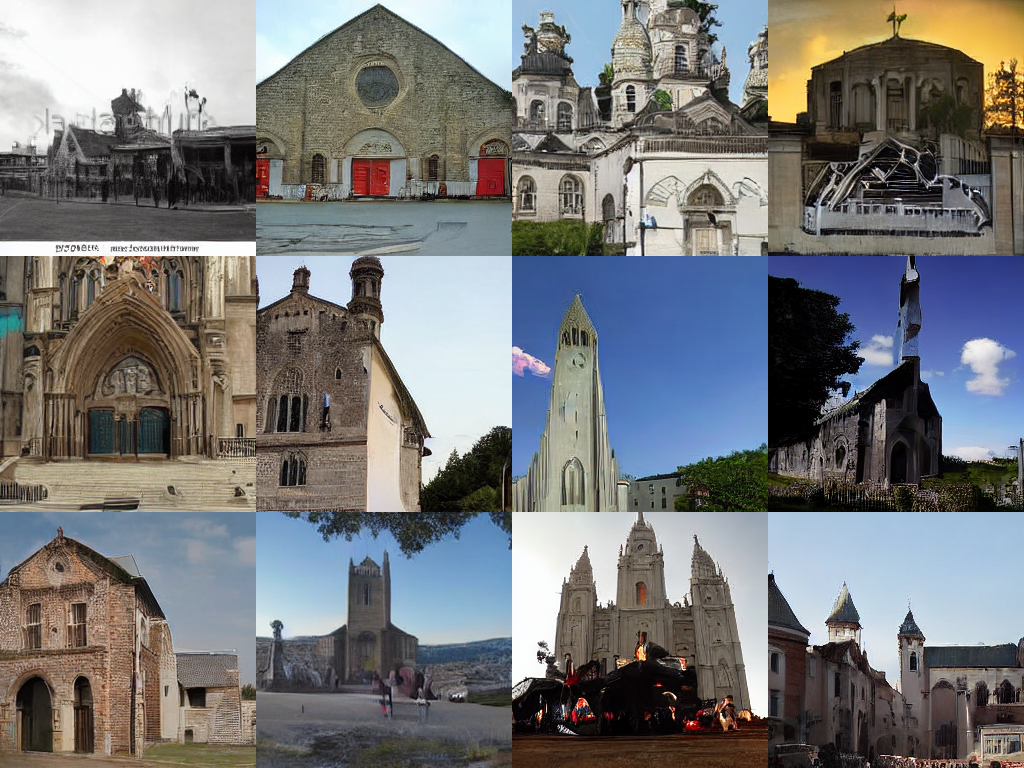} & 
        \includegraphics[width=0.49\textwidth]{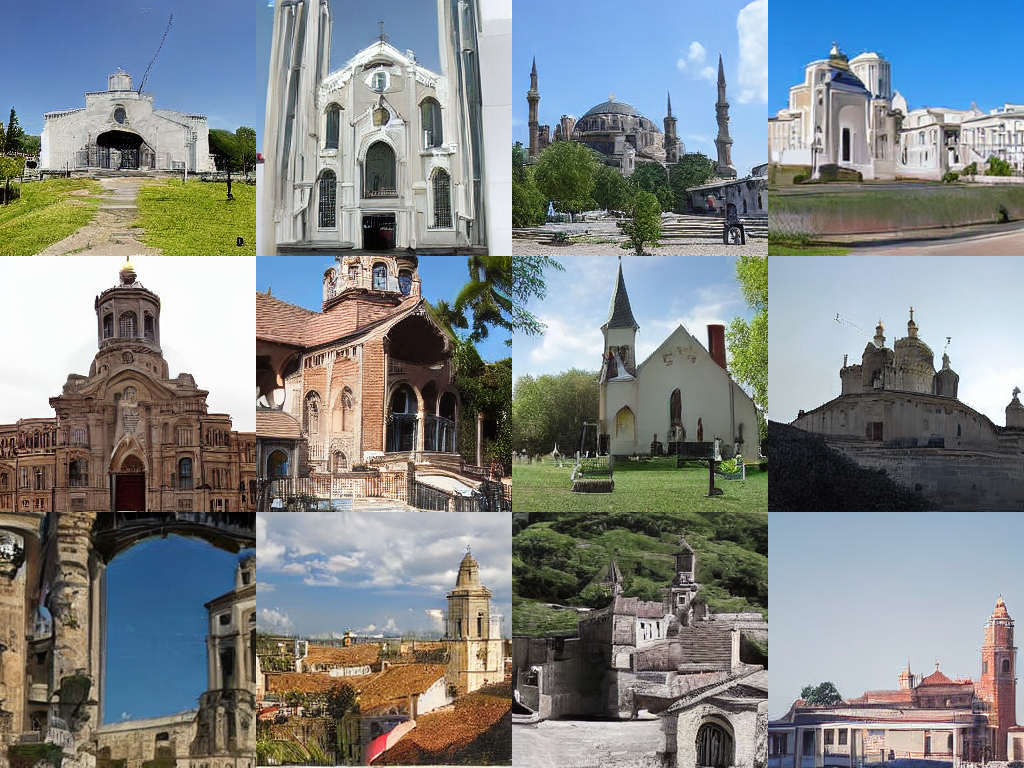} \\
        Baseline + S-PNDM (NFEs$=25$) & OGDM + S-PNDM (NFEs$=25$) \\
        (FID: 8.41, recall: 0.470) & (FID: 7.69, recall: 0.480) \\
        \includegraphics[width=0.49\textwidth]{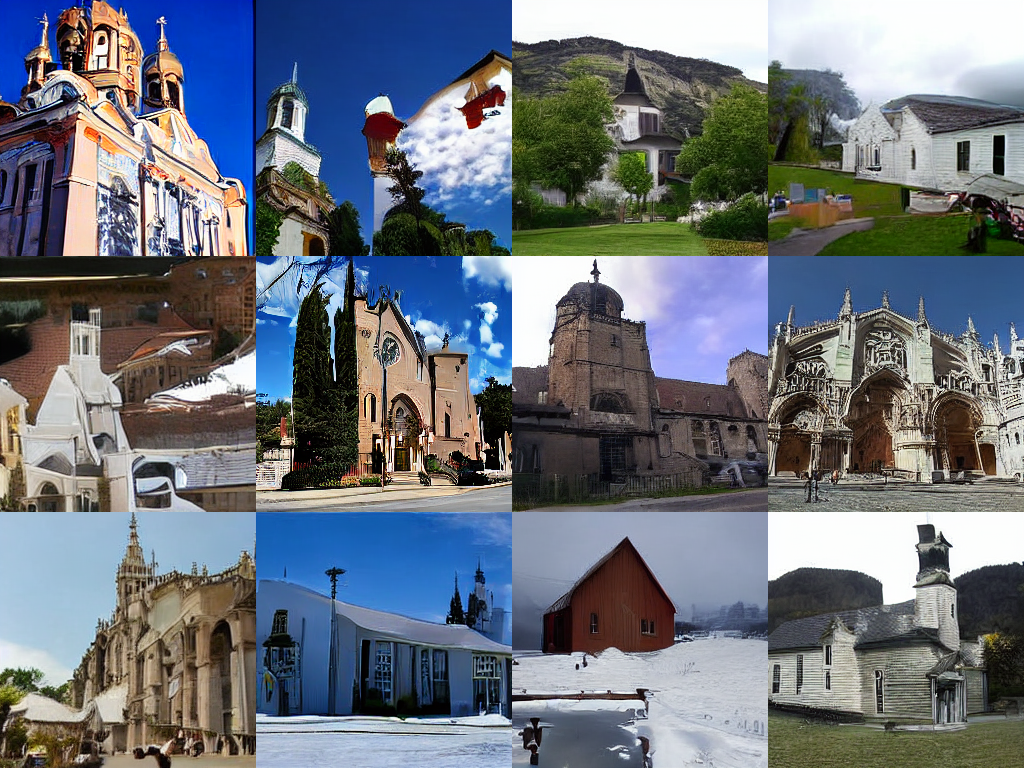} & 
        \includegraphics[width=0.49\textwidth]{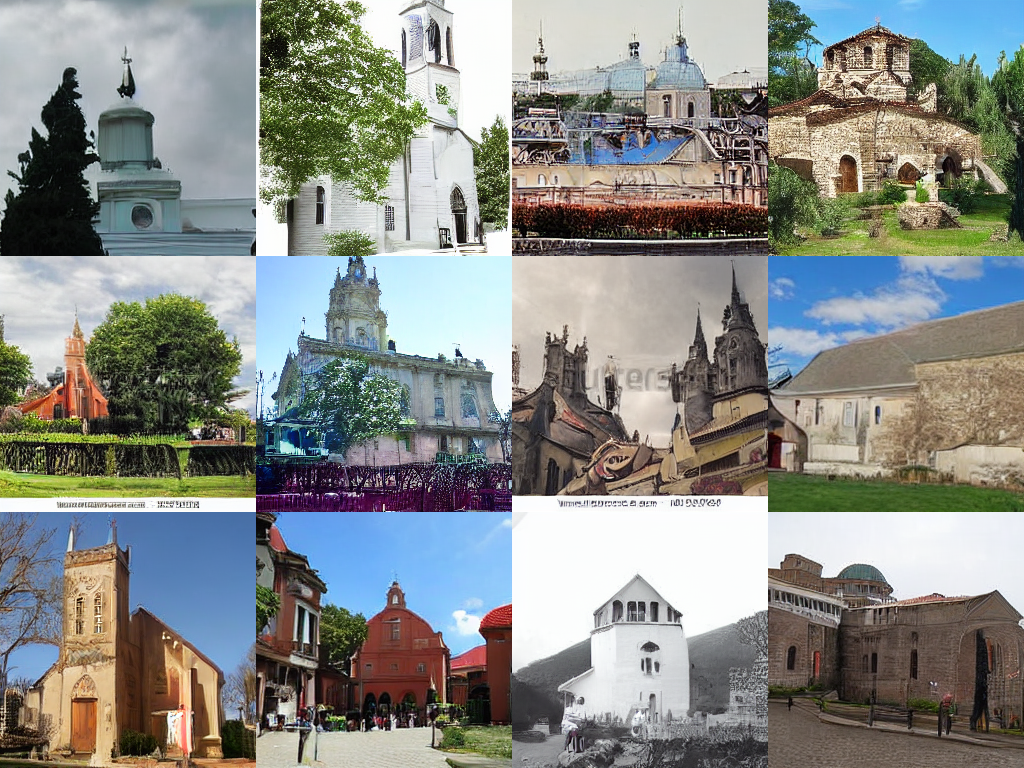} \\
        Baseline + F-PNDM (NFEs$=25$) & OGDM + F-PNDM (NFEs$=25$) \\
        (FID: 9.04, recall: 0.474) & (FID: 8.24, recall: 0.481)
    \end{tabular}
    \caption{
        Qualitative results on LSUN Church dataset with the LDM backbone using Euler method (top), S-PNDM (middle) and F-PNDM (bottom) with NFEs=$25$.
    }
    \label{fig:qual_lsun_church_nfe25}  
\end{figure*}
\clearpage
\subsection{Nearest neighborhoods}
\label{app:nn}

\end{document}